\documentclass[sigconf]{acmart}
\AtBeginDocument{%
  }

\copyrightyear{2026}
\acmYear{2026}
\setcopyright{cc}
\setcctype{by-nc-nd}
\acmConference[KDD '26]{Proceedings of the 32nd ACM SIGKDD Conference on Knowledge Discovery and Data Mining V.2}{August 09--13, 2026}{Jeju Island, Republic of Korea}
\acmBooktitle{Proceedings of the 32nd ACM SIGKDD Conference on Knowledge Discovery and Data Mining V.2 (KDD '26), August 09--13, 2026, Jeju Island, Republic of Korea}
\acmDOI{10.1145/3770855.3817591}
\acmISBN{979-8-4007-2259-2/2026/08}

\usepackage{graphicx}
\usepackage{booktabs}
\usepackage{multirow}
\usepackage{xcolor}
\usepackage{tikz}
\usepackage{tabularx}
\usepackage{cleveref}
\usepackage{rotating}
\usepackage{subcaption}
\usepackage{balance}

\begin{document}

\title{GraphARC: A Comprehensive Benchmark for Graph-Based Abstract Reasoning}

\author{Saku Peltonen}
\email{speltonen@ethz.ch}
\affiliation{%
  \institution{ETH Z\"urich}
  \city{Z\"urich}
  \country{Switzerland}
}

\author{August Bøgh Rønberg}
\email{ronberga@ethz.ch}
\affiliation{%
  \institution{ETH Z\"urich}
  \city{Z\"urich}
  \country{Switzerland}
}

\author{Andreas Plesner}
\email{aplesner@ethz.ch}
\affiliation{%
  \institution{ETH Z\"urich}
  \city{Z\"urich}
  \country{Switzerland}
}

\author{Roger Wattenhofer}
\email{wattenhofer@ethz.ch}
\affiliation{%
  \institution{ETH Z\"urich}
  \city{Z\"urich}
  \country{Switzerland}
}

\renewcommand{\shortauthors}{Saku Peltonen, August Bøgh Rønberg, Andreas Plesner, and Roger Wattenhofer}

\begin{abstract}
    Relational reasoning lies at the heart of intelligence, but existing benchmarks are typically confined to formats such as grids or text. We introduce \emph{GraphARC}, a benchmark for abstract reasoning on graph-structured data. GraphARC generalizes the few-shot transformation learning paradigm of the Abstraction and Reasoning Corpus (ARC). Each task requires inferring a transformation rule from a few input-output pairs and applying it to a new test graph, covering local, global, and hierarchical graph transformations. Unlike grid-based ARC, GraphARC instances can be generated at scale across diverse graph families and sizes, enabling systematic evaluation of generalization abilities. We evaluate state-of-the-art language models on GraphARC and observe clear limitations. Models can answer questions about graph properties but often fail to solve the full graph transformation task, revealing a comprehension-execution gap. Performance further degrades on larger instances, exposing scaling barriers. More broadly, by combining aspects of node classification, link prediction, and graph generation within a single framework, GraphARC provides a promising testbed for future graph foundation models.
\end{abstract}

\begin{CCSXML}
    <ccs2012>
        <concept>
           <concept_id>10002950.10003624.10003633</concept_id>
           <concept_desc>Mathematics of computing~Graph theory</concept_desc>
           <concept_significance>500</concept_significance>
           </concept>
        <concept>
           <concept_id>10010147.10010257.10010293.10010297</concept_id>
           <concept_desc>Computing methodologies~Logical and relational learning</concept_desc>
           <concept_significance>500</concept_significance>
           </concept>
        <concept>
           <concept_id>10010147.10010257.10010293.10010314</concept_id>
           <concept_desc>Computing methodologies~Rule learning</concept_desc>
           <concept_significance>500</concept_significance>
           </concept>
        <concept>
           <concept_id>10010147.10010178.10010179</concept_id>
           <concept_desc>Computing methodologies~Natural language processing</concept_desc>
           <concept_significance>300</concept_significance>
           </concept>
    </ccs2012>
\end{CCSXML}
    
\ccsdesc[500]{Mathematics of computing~Graph theory}
\ccsdesc[500]{Computing methodologies~Logical and relational learning}
\ccsdesc[500]{Computing methodologies~Rule learning}
\ccsdesc[300]{Computing methodologies~Natural language processing}

\keywords{few-shot, abstract reasoning, graph, scalability, benchmark, ARC, reasoning models, compositional generalization}


\maketitle

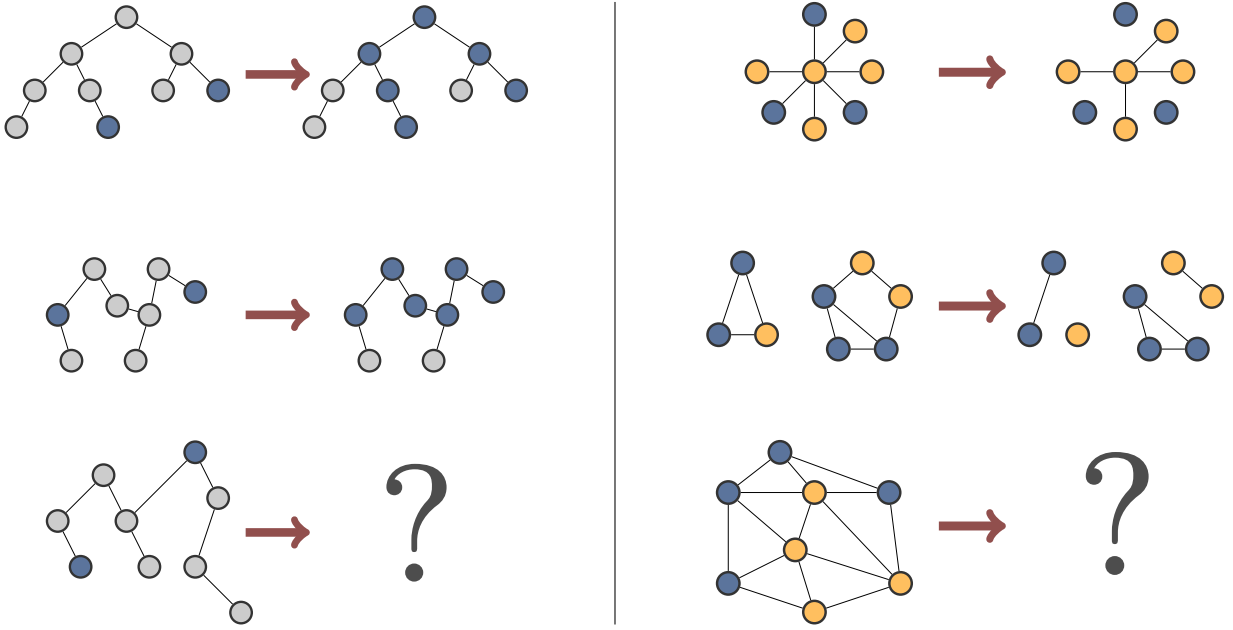
\begin{figure*}[t]
    \centering
    \resizebox{0.4\linewidth}{!}{\definecolor{myred}{RGB}{145,79,77}
\definecolor{myblue}{RGB}{73,100,145}   

\begin{tikzpicture}[
    gray_node/.style={circle, fill=gray!40, draw=black!80, line width=1.1pt, minimum size=10pt},
    blue_node/.style={circle, fill=myblue!90, draw=black!80, line width=1.1pt, minimum size=10pt},
    scale=0.75
]

% TOP LEFT GRAPH
\begin{scope}[shift={(0,10)},scale=0.8]
    % Tree structure - carefully positioned
    \node[gray_node] (a1) at (0,2.5) {};          % root
    \node[gray_node] (a2) at (-1.5,1.5) {};       % left child
    \node[gray_node] (a3) at (1.5,1.5) {};        % right child
    \node[gray_node] (a4) at (-2.5,0.5) {};       % leftmost leaf
    \node[gray_node] (a5) at (-1,0.5) {};         % left-middle leaf
    \node[gray_node] (a6) at (1,0.5) {};          % right-middle leaf  
    \node[blue_node] (a7) at (2.5,0.5) {};        % rightmost leaf
    \node[gray_node] (a8) at (-3,-0.5) {};        % bottom left
    \node[blue_node] (a9) at (-0.5,-0.5) {};      % bottom middle
    
    % Edges
    \draw (a1) -- (a2) -- (a4) -- (a8);
    \draw (a2) -- (a5) -- (a9);
    \draw (a1) -- (a3) -- (a6);
    \draw (a3) -- (a7);
\end{scope}

% TOP MIDDLE GRAPH  
\begin{scope}[shift={(0,5)}]
    % \node[gray_node] (b1) at (0,2.5) {};          % top
    \node[gray_node] (b2) at (-0.7,1.5) {};       % left-top
    \node[gray_node] (b3) at (0.7,1.5) {};        % right-top
    \node[blue_node] (b4) at (-1.5,0.5) {};       % left
    \node[gray_node] (b5) at (-0.2,0.7) {};       % center-left
    \node[gray_node] (b6) at (0.5,0.5) {};        % center-right
    \node[blue_node] (b7) at (1.5,1) {};          % right
    \node[gray_node] (b8) at (-1.2,-0.5) {};      % bottom-left
    \node[gray_node] (b9) at (0.2,-0.5) {};       % bottom-right
    
    % Edges - more complex connectivity
    \draw (b2) -- (b4) -- (b8);
    \draw (b3) -- (b6) -- (b9);
    \draw (b2) -- (b5);
    \draw (b3) -- (b7);
    \draw (b5) -- (b6);
\end{scope}

% TOP RIGHT GRAPH
\begin{scope}[shift={(0,0)}]
    \node[gray_node] (c1) at (-0.5,2) {};         % left top
    \node[blue_node] (c2) at (1.5,2.5) {};        % right top (blue)
    \node[gray_node] (c3) at (-1.5,1) {};         % left branch
    \node[gray_node] (c4) at (0,1) {};            % center
    \node[gray_node] (c5) at (2,1.5) {};          % right branch
    \node[blue_node] (c6) at (-1,0) {};           % left blue
    \node[gray_node] (c7) at (0.5,0) {};          % center bottom
    \node[gray_node] (c8) at (1.5,0) {};          % right bottom
    \node[gray_node] (c9) at (2.5,-1) {};         % far right bottom
    
    % Edges
    \draw (c1) -- (c3) -- (c6);
    \draw (c1) -- (c4) -- (c7);
    \draw (c2) -- (c4);
    \draw (c2) -- (c5) -- (c8) -- (c9);
\end{scope}

% RED ARROWS
\draw[myred, line width=4pt, ->] (2.6, 10.75) -- (4,10.75);
\draw[myred, line width=4pt, ->] (2.6, 5.5) -- (4,5.5);
\draw[myred, line width=4pt, ->] (2.6, 0.75) -- (4,0.75);

% BOTTOM LEFT GRAPH (transformed)
\begin{scope}[shift={(6.5,10)},scale=0.8]
    \node[blue_node] (d1) at (0,2.5) {};          % root - stays gray
    \node[blue_node] (d2) at (-1.5,1.5) {};       % left child - stays gray
    \node[blue_node] (d3) at (1.5,1.5) {};        % right child - stays gray
    \node[gray_node] (d4) at (-2.5,0.5) {};       % leftmost - gray to blue
    \node[blue_node] (d5) at (-1,0.5) {};         % left-middle - stays blue
    \node[gray_node] (d6) at (1,0.5) {};          % right-middle - gray to blue  
    \node[blue_node] (d7) at (2.5,0.5) {};        % rightmost - stays blue
    \node[gray_node] (d8) at (-3,-0.5) {};        % bottom left - stays gray
    \node[blue_node] (d9) at (-0.5,-0.5) {};      % bottom middle - stays gray
    
    % Edges - same structure
    \draw (d1) -- (d2) -- (d4) -- (d8);
    \draw (d2) -- (d5) -- (d9);
    \draw (d1) -- (d3) -- (d6);
    \draw (d3) -- (d7);
\end{scope}

% BOTTOM MIDDLE GRAPH (transformed)
\begin{scope}[shift={(6.5,5)}]
    % \node[gray_node] (e1) at (0,2.5) {};          % top - stays gray
    \node[blue_node] (e2) at (-0.7,1.5) {};       % left-top - gray to blue
    \node[blue_node] (e3) at (0.7,1.5) {};        % right-top - gray to blue
    \node[blue_node] (e4) at (-1.5,0.5) {};       % left - gray to blue
    \node[blue_node] (e5) at (-0.2,0.7) {};       % center-left - stays blue
    \node[blue_node] (e6) at (0.5,0.5) {};        % center-right - gray to blue
    \node[blue_node] (e7) at (1.5,1) {};          % right - stays blue
    \node[gray_node] (e8) at (-1.2,-0.5) {};      % bottom-left - stays gray
    \node[gray_node] (e9) at (0.2,-0.5) {};       % bottom-right - stays gray
    
    % Edges - same structure
    \draw (e2) -- (e4) -- (e8);
    \draw (e3) -- (e6) -- (e9);
    \draw (e2) -- (e5);
    \draw (e3) -- (e7);
    \draw (e5) -- (e6);
\end{scope}

% BOTTOM RIGHT GRAPH (transformed)
\begin{scope}[shift={(5.5,-0.3)},scale=0.03]
%     \node[gray_node] (f1) at (-0.5,2) {};         % left top - stays gray
%     \node[blue_node] (f2) at (1.5,2.5) {};        % right top - stays blue
%     \node[gray_node] (f3) at (-1.5,1) {};         % left branch - stays gray
%     \node[gray_node] (f4) at (0,1) {};            % center - stays gray
%     \node[gray_node] (f5) at (2,1.5) {};          % right branch - stays gray
%     \node[blue_node] (f6) at (-1,0) {};           % left blue - stays blue
%     \node[gray_node] (f7) at (0.5,0) {};          % center bottom - stays gray
%     \node[gray_node] (f8) at (1.5,0) {};          % right bottom - stays gray
%     \node[gray_node] (f9) at (2.5,-1) {};         % far right bottom - stays gray
    
%     % Edges - same structure
%     \draw (f1) -- (f3) -- (f6);
%     \draw (f1) -- (f4) -- (f7);
%     \draw (f2) -- (f4);
%     \draw (f2) -- (f5) -- (f8) -- (f9);
\filldraw[fill=black!70, draw=black!70, line width=1pt] (48.656200,67.906197)..controls (48.656200,75.199203) and (43.636700,84.164101)..(26.421900,84.164101)
    ..controls (13.507800,84.164101) and (6.457030,75.796898)..(6.457030,68.144501)
    ..controls (6.457030,63.839802) and (9.683590,62.882801)..(11.476600,62.882801)
    ..controls (13.507800,62.882801) and (16.378901,64.320297)..(16.378901,67.906197)
    ..controls (16.378901,70.656197) and (14.347700,72.808601)..(11.359400,72.808601)
    ..controls (10.640600,72.808601) and (10.402300,72.808601)..(10.160200,72.687500)
    ..controls (12.793000,78.902298) and (19.726601,81.773399)..(26.062500,81.773399)
    ..controls (39.570301,81.773399) and (39.570301,73.046898)..(39.570301,68.503899)
    ..controls (39.570301,61.449200) and (37.417999,59.179699)..(35.386700,57.027302)
    ..controls (27.257799,48.300800) and (24.628901,37.179699)..(24.628901,29.886700)
    --(24.628901,24.148399)..controls (24.628901,21.996099) and (24.628901,21.519501)..(25.941401,21.519501)
    ..controls (27.257799,21.519501) and (27.257799,22.355499)..(27.257799,24.507799)
    --(27.257799,28.929701)..controls (27.257799,35.984402) and (30.128901,46.503899)..(42.203098,55.472698)
    ..controls (45.550800,57.984402) and (48.656200,61.687500)..(48.656200,67.906197)
    --cycle;
\filldraw[fill=black!70, draw=black!70, line width=1pt] (31.679701,5.859380)..controls (31.679701,8.964840) and (29.050800,11.597700)..(25.941401,11.597700)
    ..controls (22.355499,11.597700) and (20.085899,8.726560)..(20.085899,5.859380)
    ..controls (20.085899,2.269530) and (22.953100,0.000000)..(25.824200,0.000000)
    ..controls (29.171900,0.000000) and (31.679701,2.628910)..(31.679701,5.859380)
    --cycle;
\end{scope}

\end{tikzpicture}}
    \hspace{1cm}\vline\hspace{1cm}%
    \resizebox{0.4\linewidth}{!}{\definecolor{myred}{RGB}{145,79,77}
\definecolor{myblue}{RGB}{73,100,145}
\definecolor{myorange}{RGB}{255, 184, 77}   

\begin{tikzpicture}[
    orange_node/.style={circle, fill=myorange!90, draw=black!80, line width=1.1pt, minimum size=10pt},
    gray_node/.style={circle, fill=myorange!60, draw=black!80, line width=1.1pt, minimum size=10pt},
    blue_node/.style={circle, fill=myblue!90, draw=black!80, line width=1.1pt, minimum size=10pt},
    node/.style={circle, fill=gray!60, draw=black, line width=1.1pt, minimum size=10pt},
    scale=0.75
]

% GRAPH 2: Star Graph (one central hub)
\begin{scope}[shift={(0,10)}]
    \node[orange_node] (b1) at (0,0) {};      % center
    \node[blue_node] (b2) at (0,1.2) {};    % top
    \node[orange_node] (b3) at (0.85,0.85) {};  % top-right
    \node[orange_node] (b4) at (1.2,0) {};    % right
    \node[blue_node] (b5) at (0.85,-0.85) {}; % bottom-right
    \node[orange_node] (b6) at (0,-1.2) {};   % bottom
    \node[blue_node] (b7) at (-0.85,-0.85) {};% bottom-left
    \node[orange_node] (b8) at (-1.2,0) {};   % left
    
    % Star connections - all through center
    \draw (b1) -- (b2);
    \draw (b1) -- (b3);
    \draw (b1) -- (b4);
    \draw (b1) -- (b5);
    \draw (b1) -- (b6);
    \draw (b1) -- (b7);
    \draw (b1) -- (b8);
\end{scope}

% GRAPH 4: Two Separate Components
\begin{scope}[shift={(0,5)}]
    % Component 1 (triangle)
    \node[blue_node] (d1) at (-1.5,1) {};
    \node[blue_node] (d2) at (-2,-0.5) {};
    \node[orange_node] (d3) at (-1,-0.5) {};
    
    % Component 2 (pentagon-like)
    \node[orange_node] (d4) at (1,1) {};
    \node[blue_node] (d5) at (0.2,0.3) {};
    \node[blue_node] (d6) at (0.5,-0.8) {};
    \node[blue_node] (d7) at (1.5,-0.8) {};
    \node[orange_node] (d8) at (1.8,0.3) {};
    
    % Component 1 connections
    \draw (d1) -- (d2) -- (d3) -- (d1);
    
    % Component 2 connections
    \draw (d4) -- (d5) -- (d6) -- (d7) -- (d8) -- (d4);
    \draw (d5) -- (d7);
\end{scope}

\begin{scope}[shift={(0,0)}]
    \node[orange_node] (a1) at (0,1.2) {};
    \node[blue_node] (a2) at (-1.8,1.2) {};
    \node[blue_node] (a3) at (-1.8,-0.7) {};
    \node[orange_node] (a4) at (0,-1.3) {};
    \node[orange_node] (a5) at (1.8,-0.7) {};
    \node[blue_node] (a6) at (1.56,1.2) {};
    \node[orange_node] (a7) at (-0.4,-0) {};
    \node[blue_node] (a8) at (-0.72,2.04) {};
    
    % Dense connections - almost complete graph
    \draw (a2) -- (a8) -- (a1) -- (a2);
    \draw (a1) -- (a6) -- (a5) -- (a4) -- (a7) -- (a2) -- (a3) -- (a4);
    \draw (a8) -- (a6);
    \draw (a3) -- (a7) -- (a1);
    \draw (a1) -- (a5) -- (a7);
\end{scope}

% RED ARROWS
\draw[myred, line width=4pt, ->] (2.6, 9.99) -- (4,9.99);
\draw[myred, line width=4pt, ->] (2.6, 5.1) -- (4,5.1);
\draw[myred, line width=4pt, ->] (2.6, 0.5) -- (4,0.5);

\begin{scope}[shift={(6.5,10)}]
    \node[orange_node] (b1) at (0,0) {};      % center
    \node[blue_node] (b2) at (0,1.2) {};    % top
    \node[orange_node] (b3) at (0.85,0.85) {};  % top-right
    \node[orange_node] (b4) at (1.2,0) {};    % right
    \node[blue_node] (b5) at (0.85,-0.85) {}; % bottom-right
    \node[orange_node] (b6) at (0,-1.2) {};   % bottom
    \node[blue_node] (b7) at (-0.85,-0.85) {};% bottom-left
    \node[orange_node] (b8) at (-1.2,0) {};   % left
    
    % Star connections - all through center
    %\draw (b1) -- (b2);
    \draw (b1) -- (b3);
    \draw (b1) -- (b4);
    %\draw (b1) -- (b5);
    \draw (b1) -- (b6);
    %\draw (b1) -- (b7);
    \draw (b1) -- (b8);
\end{scope}

\begin{scope}[shift={(6.5,5)}]
    % Component 1 (triangle)
    \node[blue_node] (d1) at (-1.5,1) {};
    \node[blue_node] (d2) at (-2,-0.5) {};
    \node[orange_node] (d3) at (-1,-0.5) {};
    
    % Component 2 (pentagon-like)
    \node[orange_node] (d4) at (1,1) {};
    \node[blue_node] (d5) at (0.2,0.3) {};
    \node[blue_node] (d6) at (0.5,-0.8) {};
    \node[blue_node] (d7) at (1.5,-0.8) {};
    \node[orange_node] (d8) at (1.8,0.3) {};
    
    % Component 1 connections
    \draw (d1) -- (d2);
    
    % Component 2 connections
    \draw (d5) -- (d6) -- (d7);
    \draw (d8) -- (d4);
    \draw (d5) -- (d7);
\end{scope}

% BOTTOM RIGHT GRAPH (transformed)
\begin{scope}[shift={(5.5,-0.5)},scale=0.03]
%     \node[gray_node] (f1) at (-0.5,2) {};         % left top - stays gray
%     \node[blue_node] (f2) at (1.5,2.5) {};        % right top - stays blue
%     \node[gray_node] (f3) at (-1.5,1) {};         % left branch - stays gray
%     \node[gray_node] (f4) at (0,1) {};            % center - stays gray
%     \node[gray_node] (f5) at (2,1.5) {};          % right branch - stays gray
%     \node[blue_node] (f6) at (-1,0) {};           % left blue - stays blue
%     \node[gray_node] (f7) at (0.5,0) {};          % center bottom - stays gray
%     \node[gray_node] (f8) at (1.5,0) {};          % right bottom - stays gray
%     \node[gray_node] (f9) at (2.5,-1) {};         % far right bottom - stays gray
    
%     % Edges - same structure
%     \draw (f1) -- (f3) -- (f6);
%     \draw (f1) -- (f4) -- (f7);
%     \draw (f2) -- (f4);
%     \draw (f2) -- (f5) -- (f8) -- (f9);
\filldraw[fill=black!70, draw=black!70, line width=1pt] (48.656200,67.906197)..controls (48.656200,75.199203) and (43.636700,84.164101)..(26.421900,84.164101)
    ..controls (13.507800,84.164101) and (6.457030,75.796898)..(6.457030,68.144501)
    ..controls (6.457030,63.839802) and (9.683590,62.882801)..(11.476600,62.882801)
    ..controls (13.507800,62.882801) and (16.378901,64.320297)..(16.378901,67.906197)
    ..controls (16.378901,70.656197) and (14.347700,72.808601)..(11.359400,72.808601)
    ..controls (10.640600,72.808601) and (10.402300,72.808601)..(10.160200,72.687500)
    ..controls (12.793000,78.902298) and (19.726601,81.773399)..(26.062500,81.773399)
    ..controls (39.570301,81.773399) and (39.570301,73.046898)..(39.570301,68.503899)
    ..controls (39.570301,61.449200) and (37.417999,59.179699)..(35.386700,57.027302)
    ..controls (27.257799,48.300800) and (24.628901,37.179699)..(24.628901,29.886700)
    --(24.628901,24.148399)..controls (24.628901,21.996099) and (24.628901,21.519501)..(25.941401,21.519501)
    ..controls (27.257799,21.519501) and (27.257799,22.355499)..(27.257799,24.507799)
    --(27.257799,28.929701)..controls (27.257799,35.984402) and (30.128901,46.503899)..(42.203098,55.472698)
    ..controls (45.550800,57.984402) and (48.656200,61.687500)..(48.656200,67.906197)
    --cycle;
\filldraw[fill=black!70, draw=black!70, line width=1pt] (31.679701,5.859380)..controls (31.679701,8.964840) and (29.050800,11.597700)..(25.941401,11.597700)
    ..controls (22.355499,11.597700) and (20.085899,8.726560)..(20.085899,5.859380)
    ..controls (20.085899,2.269530) and (22.953100,0.000000)..(25.824200,0.000000)
    ..controls (29.171900,0.000000) and (31.679701,2.628910)..(31.679701,5.859380)
    --cycle;
\end{scope}

\end{tikzpicture}}

    \caption{Two examples of GraphARC tasks demonstrating few-shot abstract reasoning on graphs. Each task presents two input-output graph pairs (top row and middle row) that illustrate a transformation rule, followed by a test input graph where the learned rule must be applied. In the example on the left-hand side, the transformation rule colors the shortest path connecting two colored nodes (shown in blue). This task is constrained to tree structures to ensure a unique shortest path between any two nodes. The example on the right-hand side shows a task where edges between nodes of different colors are removed.}
    \label{fig:example_task}
\end{figure*}

\section{Introduction}

Relational reasoning—the ability to perceive and reason about relationships between objects—is a core aspect of intelligence \citep{hummel2003symbolic, HALFORD2010497}. This capacity underlies many forms of higher cognition: we use it to appreciate analogies across different domains \citep{holyoak201213}, to understand and learn language \citep{pinker1998words}, and generally, to apply abstract rules to novel situations \citep{smith1992case}. Achieving this kind of generalization is a central challenge for artificial intelligence \citep{chollet2019measure, lake2017building}.

The Abstraction and Reasoning Corpus (ARC) \citep{chollet2019measure} is a widely recognized benchmark for evaluating abstract reasoning in AI. ARC consists of grid-based visual puzzles that require systems to learn and apply transformation rules based on a few examples. 
While ARC puzzles are grid-based, many of the underlying rules are relational—grouping identical objects, replicating subpatterns, or propagating attributes to neighboring objects. 
To capture such a structure more directly, we propose to use graphs as a more general representation that is not tied to a particular spatial layout.

Inspired by ARC, we introduce \emph{GraphARC}, a benchmark for few-shot abstract reasoning on graphs. Each task includes 2-3 input-output graph pairs demonstrating a transformation rule, and a test input where the rule should be applied. See \Cref{fig:example_task} for an example. The transformations are based on fundamental graph primitives: local structure (degree, neighborhoods, cliques), reachability (connected components, isolated nodes), or hierarchical relations (closest common ancestor in a tree). The transformations can be color-based (changing the color of some nodes) and structural modifications, adding or removing nodes and edges. The instances are automatically generated across diverse graph families and sizes, providing a virtually unlimited supply of instances. This setup allows testing whether models can generalize the same transformation across graphs of varying size and structure.

GraphARC combines elements of traditional graph learning tasks, including node classification \citep{kipf2016semi}, graph classification \citep{ying2018hierarchical}, link prediction \citep{zhang2018link}, and graph generation \citep{simonovsky2018graphvae} within a single framework. 
Given the absence of broadly applicable Graph Foundation Models (GFMs) \citep{liu2025graph, wang2025graph}, we focus on evaluating Large Language Models (LLMs) that can process textual representations of graphs. To this end, we systematically test a range of encoding schemes and prompting choices. We also design two complementary evaluation pipelines for LLMs: one that measures performance on full graph transformation tasks, and another that uses targeted questions about properties of the input and output graph. This allows us to disentangle a model's ability to understand the transformation from its ability to execute it. 

Concretely, GraphARC provides 
\begin{enumerate}
    \item a scalable task generation framework that produces diverse graph transformation challenges across multiple graph families at arbitrary scales,
    \item a comprehensive evaluation methodology for LLMs, and
    \item an extensive analysis of current LLM performance, revealing significant gaps in structural understanding and identifying key failure modes that highlight directions for future research.
\end{enumerate}
Our code is available on Github.\footnote{\url{https://github.com/sakupeltonen/graph-arc}}

\section{Related work}

\paragraph{The Abstraction and Reasoning Corpus.}

The ARC challenge, introduced by \citet{chollet2019measure}, established a paradigm for few-shot abstract reasoning in which systems infer transformation rules from minimal input–output examples and apply them to new cases. Despite intensive effort, performance hovered near 33\% for years before recent breakthroughs: top open-source entries in the ARC Prize combined neural reasoning with programmatic search to reach roughly 53\% accuracy. At the same time, OpenAI's o3 attained 75.7\% on ARC-AGI-1 \citep{arcprize2024progress,arcprize2024}. However, the release of ARC-AGI-2 \citep{chollet2025arc} in March 2025 largely reset the field, with frontier systems scoring about 16\% versus human performance near 60\%, underscoring that robust abstract reasoning remains open. Crucially for our setting, leading ARC solvers are tailored to grid-based vision puzzles and often rely on specialized pipelines or substantial compute, making direct application to GraphARC's relational, graph-structured I/O unnecessary or impractical. Accordingly, we evaluate general-purpose LLMs on GraphARC's textual graph encodings to probe abstract relational reasoning without bespoke ARC-specific machinery.

The impact of ARC has led to several extensions exploring different modalities and task formulations. \citep{lee2024llms} created 1D-ARC tasks specifically for language model evaluation. \citet{assouel2022object} developed Arith-MNIST, containing reasoning tasks where models must infer arithmetic programs applied to colored digits. In this challenge, the output is a single digit containing the answer, rather than a transformed grid. 

\paragraph{Large Language Models and Graph Reasoning.}
Recent work has explored applying LLMs to graph reasoning tasks with mixed results. 
\citet{fatemi2023talk} explores various textual representations and their impact on LLM performance across different reasoning tasks.
\citet{wang2023can} examines whether language models can solve graph problems such as connectivity, cycle existence, and bipartite matching, when graphs are described in text. 

\citet{dai2024large} evaluate how LLMs understand graph patterns through tasks such as detection, translation, and modification, with patterns specified in natural language or as edge lists. This work is arguably the most similar to our work, but it focuses on reasoning about predefined motifs, whereas we target few-shot learning and the application of general graph transformations.
Beyond inputting graphs in text, \citet{zhao2023gimlet} introduces GIMLET, using a customized positional encoding to integrate language models with graph-structured data for molecule property prediction. 
\citet{sanford2024understanding} investigates how well transformers can solve graph-based reasoning problems at various model sizes. While they do not use natural language to represent the graphs, they do show that transformers can solve graph problems they are trained on. 
See \citep{jin2024large} for a comprehensive survey on LLMs for graph problems.

Chain-of-thought prompting (or test-time compute) has emerged as an effective way to improve reasoning abilities in language models \citep{wei2022chain,snell2025scaling,mirtaheri2025let}. We will investigate this by using closed- and open-source models with reasoning abilities, such as OpenAI's o3-mini and DeepSeek's R1. 

\paragraph{Graph Foundation Models.} GFMs are an emerging class of models aimed at general-purpose graph reasoning. These approaches seek to combine GNN-style structural inductive biases with the broad knowledge and few-shot learning of foundation models \citep{liu2023towards}. Initial efforts have targeted specific domains, such as knowledge graphs \citep{galkin2024ultraquery} and molecular graphs \citep{mole}. 
See \citep{liu2025graph} for a survey outlining the opportunities and challenges. 

Notably, Google Research has announced a relational GFM that treats databases as graphs and generalizes to unseen tables, reporting up to 40x precision gains on tasks like spam detection. Despite rapid progress, current GFMs largely target standard supervised objectives (node/graph classification, link prediction) \citep{liu2023towards,liu2025graph,wang2025graph}, and cannot be straightforwardly adapted to solve ARC style questions. 
Accordingly, we do not benchmark GFMs here; instead, GraphARC serves as a complementary proving ground for future GFMs that claim abstract, few-shot graph transformation capabilities.

\section{GraphARC}

\begin{table}[ht!]
    \centering
    \caption{GraphARC transformations. The tasks are grouped into color-based transformations and structural transformations.}
    \label{tab:tasks}
    \small
    \begin{tabular}{p{0.33\linewidth}p{0.6\linewidth}}
    \toprule
    \textbf{Task} & \textbf{Transformation rule} \\
    \midrule
    \multicolumn{2}{l}{\textit{Color-based transformations}} \\
    \midrule
    \texttt{colorDegreeX} & Color all nodes with degree exactly $X$. \\
    \texttt{colorMaxDegree} & Color all nodes with maximum degree. \\
    \texttt{colorMinDegree} & Color all nodes with minimum degree. \\
    \texttt{colorInternal} & Color all non-leaf nodes, i.e., nodes with degree greater than $1$. \\
    \texttt{colorNeighbors} & Color all neighbors of nodes with a specified color. \\
    \texttt{colorPath} & Color all nodes on the shortest path between two colored nodes. \\
    \texttt{colorComponents} & Color nodes according to connected component membership. \\
    \texttt{colorDistanceAtLeast2} & Color nodes at distance at least $2$ from marked nodes. \\
    \texttt{colorEquidistant} & Color nodes that are equidistant from two colored nodes. \\
    \texttt{bipartitionCompletion} & Complete a bipartite coloring from seed nodes. \\
    \midrule
    \multicolumn{2}{l}{\textit{Structural transformations}} \\
    \midrule
    \texttt{addHub} & Add a new node connected to all existing nodes. \\
    \texttt{edgeToNode} & Replace each edge by a new intermediate node. \\
    \texttt{removeDegreeX} & Remove all nodes with degree exactly $X$. \\
    \texttt{blueSubgraph} & Return the subgraph induced by nodes of a given color. \\
    \texttt{mergeAtBlue} & Merge two connected components at their colored nodes. \\
    \texttt{complementGraph} & Return the complement graph. \\
    \texttt{removeSameColorEdges} & Remove edges between nodes of the same color. \\
    \bottomrule
    \end{tabular}
\end{table}

\begin{figure*}[t!]
    \centering
    \includegraphics[width=\linewidth]{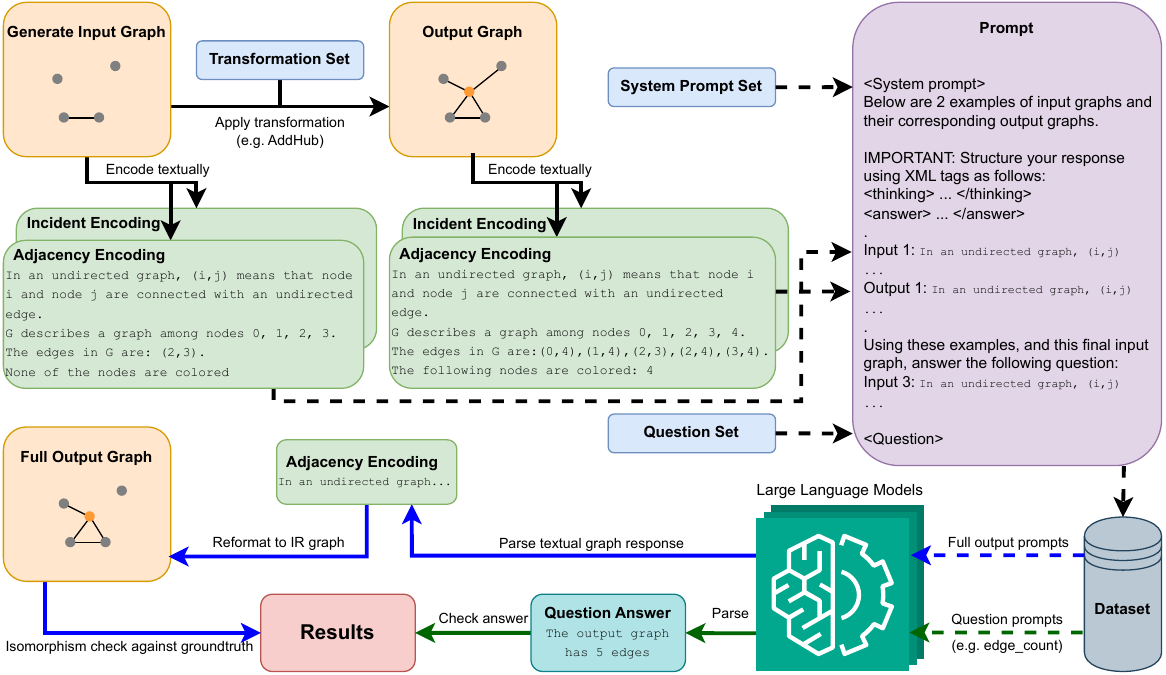}
    \caption{The GraphARC task generation and evaluation pipeline for LLMs. A task begins with a formal rule (e.g., ``color all degree-3 nodes blue"), which is used to generate and textually encode a set of input-output graph examples. The pipeline then diverges into two distinct evaluation pathways: 
    1) The \textbf{Full Output Path (blue arrows)} requires the model to generate the complete transformed graph, testing both its ability to infer the rule and execute it correctly.
    2) The \textbf{Question-Based Path (green arrows)} isolates comprehension by asking targeted questions about the input graph and the inferred output properties, disentangling understanding from execution.
    Solid arrows represent stages where analysis or transformations occur, while dashed arrows indicate the flow of data without modification.}
    \label{fig:task_pipeline}
\end{figure*}

\subsection{Benchmark Definition}

A GraphARC task consists of 2-3 input-output graph pairs that demonstrate a transformation rule. This is followed by a test input graph, where the learned rule must be applied. See \Cref{fig:example_task} for an example task. 

Formally, graphs are represented as a tuple $G = (V, E)$ where nodes have unique integer identifiers from $0$ to $|V|-1$. Node IDs remain consistent within each task instance. Each node $v \in V$ carries a color attribute, with grey as the default color. We assume the graphs to be undirected. 

GraphARC contains 21 distinct transformations. A task is an instance of a transformation. We can generate multiple tasks per transformation by varying transformation parameters, size and family of the graphs, and the number of examples. Furthermore, by varying the test graph size, we can evaluate how well models generalize transformations to larger instances. Each transformation is designed to test a specific aspect of graph reasoning, such as local structure (e.g., degree-based coloring), reachability (connected components, paths), or hierarchical relations (e.g., common ancestors in trees). 

\paragraph{Transformations.} See Table~\ref{tab:tasks} for the list of transformations. We organize GraphARC transformations into two main categories: color-based transformations and structure modification transformations. Color-based tasks modify node colors based on structural properties, whereas structure modification tasks alter the graph topology. A complete task specification includes: 
\begin{itemize}
    \item \textit{Transformation rule:} The operation to be performed (e.g., ``color all degree-3 nodes blue").
    \item \textit{Required Properties:} A list of preconditions the input graph must satisfy for the transformation to be meaningful. For instance, the ``colorDegree3" task requires the input graph to have at least one node of degree 3. This prevents the generation of trivial examples where the transformation has no effect.
    \item \textit{Parameters:} Configurable elements of the transformation, such as the target color or degree number.
    \item \textit{Pre-transformation Steps:} Initial modifications to the input graph to set up the reasoning problem. For example, the ``colorNeighbors" task first colors a random  node orange, and the model must then infer the rule to color all of its neighbors blue.
\end{itemize}

\paragraph{Graph Generation and Validation}
We generate graphs from multiple families, including Erdős-Rényi, Watts-Strogatz, trees, star, bipartite, and multi-component. This ensures diversity in structural properties, and also allows us to test whether models can learn transformations that generalize across different graph types. 

The number of examples and their size is varied according to size patterns (see Appendix \ref{app: patterns}). In the main dataset, we use small graphs with 5 to 15 nodes in both the examples and the test input. In the scaling dataset, we fix example graphs at 10 nodes, while testing on larger graphs up to 250 nodes. This allows us to evaluate how well models can generalize transformations to larger instances. 

To ensure that generated tasks are valid and solvable, we implement a property-validation system that checks whether the generated graphs satisfy necessary conditions for the transformation to be applied. It is also important that the combination of input examples should uniquely determine the simplest consistent transformation. Our system uses a three-state framework (True/False/Maybe) indicating whether a generator guarantees, precludes, or conditionally satisfies the required properties, enabling efficient rejection of invalid instances. We also set the input graph size to be large enough to make the probability of multiple valid, simple transformations negligible. Full details appear in Appendix~\ref{app:generation_validation}.

\subsection{LLM Evaluation Pipeline}
\label{sec:llm_evaluation}

\Cref{fig:task_pipeline} illustrates our task generation and evaluation pipeline. It consists of four parts. The first part consists of choosing a transformation (\textit{Stage 1}), and generating graphs and validating the examples (\textit{Stage 2}), as detailed above. These two are independent of the evaluation of LLMs and are used to generate the task instances. The last two stages are specific to the evaluation of LLMs.

\paragraph{Stage 3: Encoding and Prompt Assembly} We implement two textual encoding schemes for presenting graphs to language models: adjacency list format (enumerate edges) and incidence list format (list of neighbors by node). \Cref{fig:task_pipeline} illustrates the adjacency list format and an example incidence list can be found in \Cref{app: encoding}. 

We also test the effect of different system prompts. These encourage different approaches to the task, such as a graph analyst studying examples, a graph algorithm developer analyzing patterns, or a mathematics teacher explaining the pattern. We also include a baseline with no system prompt. See Appendix \ref{app: encoding} for a full description of these variations.

To achieve a fine-grained analysis of model capabilities, a single set of generated examples is used to create prompts for two complementary evaluation paths.

\paragraph{Stage 4 (Full Output Path):} This pathway assesses a model's ability to infer the transformation rule from the examples and apply it to generate the complete output graph for a new input. The model's response is parsed to reconstruct the output graph's structure and node colors. This reconstructed graph is then compared to the ground-truth output graph. The model's output must be an exact match, verified via a graph isomorphism check (this is fast for labeled graphs).

\paragraph{Stage 4 (Question-Based Path):} This pathway is designed to disentangle a model's comprehension of graph properties from its ability to execute a full transformation. Instead of requesting the full output graph, the prompt poses a targeted question about a specific property. Crucially, we ask parallel questions about both the visible test input (e.g., How many blue nodes are in the input?") and the unseen, to-be-inferred output (e.g., How many blue nodes will be in the output?"). The model's response, typically a numerical or categorical answer, is compared against the ground-truth value, which is computed from the actual input or output graphs. This pathway allows for a more granular diagnosis of failures, helping to distinguish between a failure to parse the input, a failure to understand the transformation, or a failure to reason about the output.

\section{Experiments and Results}

\subsection{Dataset and Models}

We evaluate a range of state-of-the-art AI systems to provide comprehensive coverage of reasoning capabilities, including Qwen3 (1.7B to 32B), DeepSeek R1, OpenAI's reasoning models o1-mini, o3-mini, o4-mini, and GPT 5 (all with medium reasoning effort), and GPT 4.1-nano as a direct-answer baseline. 

There are two regimes in the dataset: (i) our \emph{main experiments}, which span a range of tasks under multiple prompts and encoding variations, and (ii) \emph{scaling experiments}, which assess performance as graph size increases. In the main experiments, we use small graphs, \(n\in\{5,10,15\}\) for both the example and test graphs. In the scaling experiments, examples are fixed at \(n=10\), while test inputs range up to \(n=250\) with \(n\in\{10,25,50,100,250\}\).

The total number of evaluations per model results from combining different encodings, prompt variants, size patterns, transformations, and question types. After filtering by task-generator compatibility, for Qwen3 models and GPT~4.1-nano, this gives around 18k evaluations each, while reasoning models—restricted to a single prompt—cover a proportionally smaller set. The scaling dataset adds 4.4k evaluations for reasoning models.

\subsection{Results}

\begin{figure*}[t]
\centering
\includegraphics[width=\linewidth]{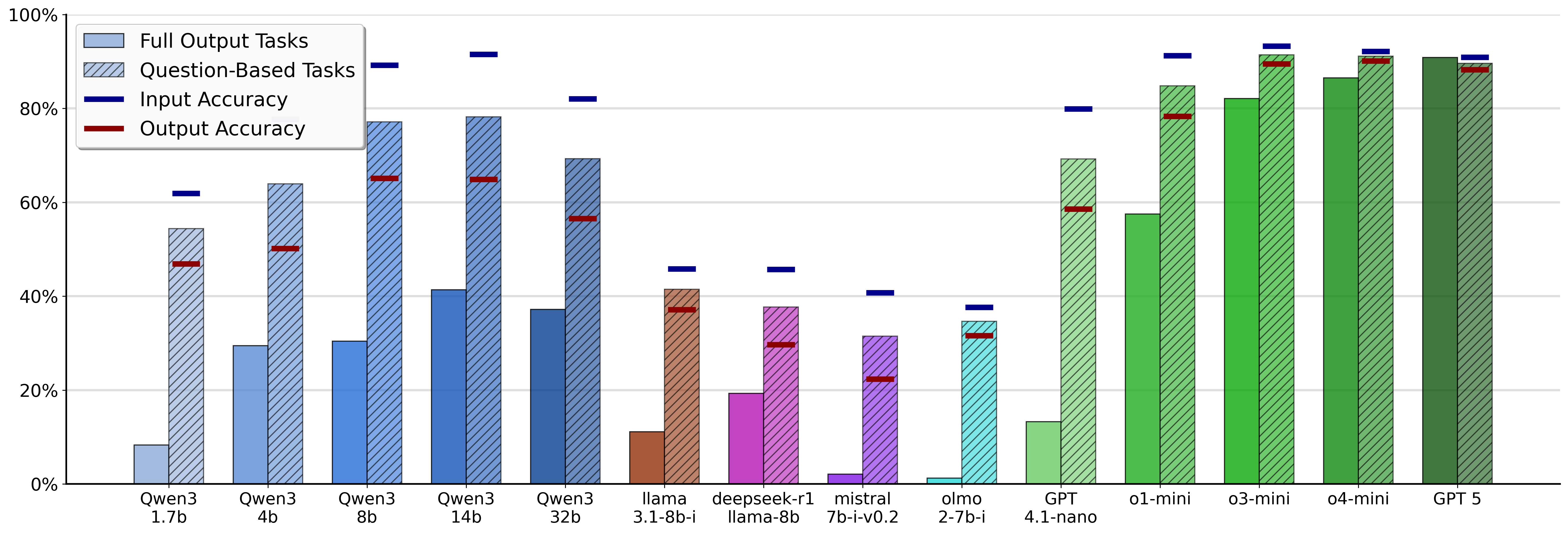}
\caption{Overall model performance on GraphARC. Bars show accuracy for full-output tasks (structured graph generation) and for question-based tasks, split into questions about the \emph{input} (visible graph) and the \emph{output} (inferred graph). For most models, we observe a comprehension--execution gap (question-based $>$ full-output) and an input--output asymmetry (input $>$ output). Reasoning models consistently outperform direct-answer baselines.}

\label{fig:overall_performance}
\end{figure*}

\begin{table*}[h]
    \centering
    \setlength{\tabcolsep}{1.0mm}
    \caption{Accuracy on full-output tasks across a subset of the tested models. Values are mean accuracies, and bold highlights the best score per row. Tasks are ordered by increasing average difficulty. Appendix \ref{app:full results} shows results and confidence intervals for all models.}
    \vspace{1mm}
    \label{tab:task_performance}
    \begin{tabular}{lccccccccccc|c}
        \toprule
        Task & \shortstack[c]{Qwen3\\14b} & \shortstack[c]{Qwen3\\32b} & \shortstack[c]{LLaMA3.1\\8b} & \shortstack[c]{DeepSeek-R1\\LLaMA-8b} & \shortstack[c]{Mistral\\7b-v0.2} & \shortstack[c]{OLMo-2\\7b} & \shortstack[c]{GPT\\4.1-nano} & \shortstack[c]{o1\\mini} & \shortstack[c]{o3\\mini} & \shortstack[c]{o4\\mini} & \shortstack[c]{GPT\\5} & Avg \\
        \midrule
        removeDegree3 & $0.02$ & $0.02$ & $0.00$ & $0.00$ & $0.00$ & $0.00$ & $0.00$ & $0.00$ & $0.25$ & $0.50$ & $\textbf{0.75}$ & $0.14$ \\
        colorDegree3 & $0.29$ & $0.06$ & $0.00$ & $0.10$ & $0.00$ & $0.00$ & $0.00$ & $0.00$ & $\textbf{0.67}$ & $0.42$ & $0.58$ & $0.19$ \\
        colorEquidistant & $0.00$ & $0.16$ & $0.00$ & $0.00$ & $0.00$ & $0.03$ & $0.00$ & $0.38$ & $0.50$ & $\textbf{0.75}$ & $\textbf{0.75}$ & $0.23$ \\
        colorDistanceAtLeast2 & $0.00$ & $0.00$ & $0.00$ & $0.00$ & $0.00$ & $0.00$ & $0.00$ & $0.08$ & $0.83$ & $\textbf{0.92}$ & $\textbf{0.92}$ & $0.25$ \\
        removeDegree2 & $0.00$ & $0.00$ & $0.00$ & $0.00$ & $0.00$ & $0.00$ & $0.00$ & $0.42$ & $0.75$ & $0.92$ & $\textbf{1.00}$ & $0.28$ \\
        colorMaxDegree & $0.52$ & $0.38$ & $0.25$ & $0.04$ & $0.05$ & $0.00$ & $0.25$ & $0.25$ & $0.33$ & $0.58$ & $\textbf{0.67}$ & $0.30$ \\
        bipartitionCompletion & $0.00$ & $0.00$ & $0.00$ & $0.00$ & $0.00$ & $0.00$ & $0.00$ & $0.75$ & $\textbf{1.00}$ & $\textbf{1.00}$ & $\textbf{1.00}$ & $0.34$ \\
        mergeAtBlue & $0.00$ & $0.12$ & $0.00$ & $0.00$ & $0.00$ & $0.00$ & $0.00$ & $0.75$ & $\textbf{1.00}$ & $\textbf{1.00}$ & $\textbf{1.00}$ & $0.35$ \\
        edgeToNode & $0.25$ & $0.33$ & $0.00$ & $0.00$ & $0.00$ & $0.00$ & $0.00$ & $0.50$ & $0.88$ & $0.94$ & $\textbf{1.00}$ & $0.35$ \\
        colorDegree2 & $0.42$ & $0.15$ & $0.00$ & $0.06$ & $0.00$ & $0.00$ & $0.04$ & $0.42$ & $0.92$ & $\textbf{1.00}$ & $\textbf{1.00}$ & $0.36$ \\
        complementGraph & $0.19$ & $0.17$ & $0.17$ & $0.16$ & $0.00$ & $0.00$ & $0.05$ & $0.69$ & $\textbf{1.00}$ & $0.94$ & $\textbf{1.00}$ & $0.40$ \\
        colorMinDegree & $0.58$ & $0.60$ & $0.02$ & $0.29$ & $0.00$ & $0.00$ & $0.12$ & $0.67$ & $0.75$ & $0.67$ & $\textbf{0.83}$ & $0.41$ \\
        removeSameColorEdges & $0.46$ & $0.48$ & $0.02$ & $0.21$ & $0.00$ & $0.00$ & $0.02$ & $0.75$ & $0.75$ & $\textbf{1.00}$ & $\textbf{1.00}$ & $0.43$ \\
        colorInternal & $0.44$ & $0.27$ & $0.23$ & $0.20$ & $0.04$ & $0.05$ & $0.25$ & $0.62$ & $\textbf{1.00}$ & $0.88$ & $\textbf{1.00}$ & $0.45$ \\
        blueSubgraph & $\textbf{0.75}$ & $0.73$ & $0.03$ & $0.56$ & $0.00$ & $0.03$ & $0.17$ & $\textbf{0.75}$ & $\textbf{0.75}$ & $\textbf{0.75}$ & $\textbf{0.75}$ & $0.48$ \\
        removeDegree1 & $0.41$ & $0.36$ & $0.19$ & $0.34$ & $0.10$ & $0.03$ & $0.19$ & $0.75$ & $\textbf{1.00}$ & $\textbf{1.00}$ & $\textbf{1.00}$ & $0.49$ \\
        colorPath & $0.41$ & $0.50$ & $0.16$ & $0.09$ & $0.03$ & $0.00$ & $0.31$ & $0.88$ & $\textbf{1.00}$ & $\textbf{1.00}$ & $\textbf{1.00}$ & $0.49$ \\
        colorComponents & $0.62$ & $0.50$ & $0.06$ & $0.44$ & $0.07$ & $0.00$ & $0.19$ & $\textbf{1.00}$ & $0.75$ & $\textbf{1.00}$ & $\textbf{1.00}$ & $0.51$ \\
        addHub & $0.66$ & $0.66$ & $0.38$ & $0.50$ & $0.06$ & $0.00$ & $0.20$ & $0.75$ & $\textbf{1.00}$ & $\textbf{1.00}$ & $\textbf{1.00}$ & $0.56$ \\
        colorNeighbors & $0.88$ & $0.88$ & $0.19$ & $0.14$ & $0.04$ & $0.03$ & $0.44$ & $0.94$ & $0.94$ & $\textbf{1.00}$ & $0.88$ & $0.58$ \\
        colorDegree1 & $0.88$ & $0.72$ & $0.25$ & $0.45$ & $0.04$ & $0.05$ & $0.27$ & $0.88$ & $\textbf{1.00}$ & $\textbf{1.00}$ & $\textbf{1.00}$ & $0.59$ \\
        \midrule
        average & $0.37$ & $0.34$ & $0.09$ & $0.17$ & $0.02$ & $0.01$ & $0.12$ & $0.58$ & $0.81$ & $0.87$ & $0.91$ & $0.39$ \\
        \bottomrule
        \end{tabular}
\end{table*}

\Cref{fig:overall_performance} reveals a stark performance hierarchy in full graph generation tasks. The Qwen3 series shows clear scaling with model size (8.3\% for 1.7B to 41.4\% for 14B), while reasoning models achieve substantially higher performance (57.5\% for o1-mini to 90.9\% for GPT 5). GPT 4.1-nano achieves only 13.3\% accuracy, performing similarly to the smallest Qwen3 model.

Models are substantially better in question-answering tasks compared to full graph generation. This gap is most pronounced in smaller models, \texttt{Qwen3 1.7B} achieves 54.4\% on questions but only 8.3\% on full outputs (6.55x difference), while \texttt{Qwen3 8B} shows 72.2\% vs 30.5\% (2.36x difference).
Even frontier models exhibit this pattern: o4-mini achieves 91.2\% on questions versus 86.5\% on full outputs. Models consistently perform better on questions about input graphs (which they can see) than output graphs (which they must infer). For example, \texttt{Qwen3 8B} has an accuracy of 89.2\% on input questions versus 65.1\% on output questions.

\paragraph{Task-Specific Performance.} The task-level performance in \Cref{tab:task_performance} shows a wide range of difficulty across tasks. Some transformations, such as \texttt{addHub}, \texttt{colorDegree1}, and \texttt{colorNeighbors}, are consistently solved with high accuracy. In contrast, others, including \texttt{removeDegree3}, \texttt{colorEquidistant}, and \texttt{colorDegree3}, remain challenging even for the strongest models. Global transformations like \texttt{bipartitionCompletion} and \texttt{mergeAtBlue} exemplify the architectural divide between model families. Only reasoning models achieve meaningful performance ($\geq75\%$), while all Qwen3 models and GPT 4.1-nano fail almost completely. See Appendix \ref{app:full results} for full results across all models.

\paragraph{Scaling Limitations.}

\begin{table}[t]
    \centering
    % \small
    \caption{Performance by size pattern (full output; aggregated across tasks). The first two columns indicate the example and test graph sizes (in nodes), and the third column gives the number of evaluated samples. Values are accuracies, with the best-performing model for each pattern in bold. The first two rows correspond to main dataset patterns; the last five rows to scaling patterns.}
    \vspace{1mm}
    \label{tab:pattern_performance}
    \begin{tabular}{lll|ccccc}
    \toprule
    \shortstack[c]{Example\\ sizes} & \shortstack[c]{Test\\ size} & $n$ & \shortstack[c]{GPT\\4.1-nano} & \shortstack[c]{o1\\mini} & \shortstack[c]{o3\\mini} & \shortstack[c]{o4\\mini} & \shortstack[c]{GPT\\5} \\
    \midrule
    5,10,15 & 15 & 1007 & $0.14$ & $0.60$ & $0.85$ & $0.90$ & $\textbf{0.95}$ \\
    5,10 & 15 & 1008 & $0.12$ & $0.56$ & $0.79$ & $0.83$ & $\textbf{0.87}$ \\
    10,10 & 10 & 224 & --- & $0.88$ & $\textbf{1.00}$ & $\textbf{1.00}$ & $\textbf{1.00}$ \\
    10,10 & 25 & 224 & --- & $0.54$ & $0.79$ & $0.96$ & $\textbf{1.00}$ \\
    10,10 & 50 & 176 & --- & $0.45$ & $0.82$ & $0.89$ & $\textbf{0.98}$ \\
    10,10 & 100 & 176 & --- & $0.34$ & $0.64$ & $0.80$ & $\textbf{0.93}$ \\
    10,10 & 250 & 176 & --- & $0.18$ & $0.41$ & $0.43$ & $\textbf{0.91}$ \\
    \bottomrule
    \end{tabular}
\end{table}

\begin{table}[t]
    \centering
    % \small
    \caption{Task performance by scaling pattern. Cells are accuracies averaged over selected models (GPT-5, o1-mini, o3-mini, and o4-mini). Column headers show the example graph sizes (before the arrow) and the test graph size (after the arrow).}
    \vspace{1mm}
    \label{tab:task_by_pattern_scaling}
    \begin{tabular}{lccccc}
    \toprule
    Task & \shortstack[c]{10,10 \\ $\to$ 10} & \shortstack[c]{10,10 \\ $\to$ 25} & \shortstack[c]{10,10 \\ $\to$ 50} & \shortstack[c]{10,10 \\ $\to$ 100} & \shortstack[c]{10,10 \\ $\to$ 250} \\
    \midrule
    removeDegree3 & $\textbf{0.96}$ & $0.67$ & $0.50$ & $0.44$ & $0.25$ \\
    removeDegree2 & $\textbf{0.92}$ & $0.75$ & $0.62$ & $0.38$ & $0.25$ \\
    bipartitionCompletion & $\textbf{0.88}$ & $0.62$ & $0.75$ & $0.62$ & $0.25$ \\
    colorDegree3 & $\textbf{1.00}$ & $0.62$ & $0.62$ & $0.56$ & $0.31$ \\
    colorDegree2 & $\textbf{1.00}$ & $0.83$ & $0.75$ & $0.62$ & $0.38$ \\
    addHub & $\textbf{1.00}$ & $0.81$ & $0.84$ & $0.66$ & $0.53$ \\
    removeDegree1 & $0.91$ & $\textbf{0.94}$ & $0.83$ & $0.88$ & $0.58$ \\
    colorComponents & $\textbf{1.00}$ & $\textbf{1.00}$ & $\textbf{1.00}$ & $\textbf{1.00}$ & $0.38$ \\
    colorDegree1 & $\textbf{1.00}$ & $0.94$ & $0.96$ & $0.79$ & $0.71$ \\
    colorPath & $\textbf{1.00}$ & $\textbf{1.00}$ & $0.88$ & $0.81$ & $0.81$ \\
    \midrule
    Average & $0.97$ & $0.82$ & $0.78$ & $0.68$ & $0.44$ \\
    \bottomrule
    \end{tabular}
\end{table}

\begin{table*}[h]
    \centering
    \small
    \caption{Performance by question type (question-based; input vs. output). Values are accuracies; bold indicates the best model per question for input and for output.}
    \vspace{1mm}
    \label{tab:questiontype_by_model_io}
    \begin{tabular}{lcccccccccc|cc}
    \toprule
     & \multicolumn{2}{c}{\shortstack[c]{Qwen3\\32b}} & \multicolumn{2}{c}{\shortstack[c]{LLaMA3.1\\8b}} & \multicolumn{2}{c}{\shortstack[c]{DeepSeek-R1\\LLaMA-8b}} & \multicolumn{2}{c}{\shortstack[c]{GPT\\4.1-nano}} & \multicolumn{2}{c}{\shortstack[c]{GPT\\5}} & \multicolumn{2}{c}{Avg.} \\
    \cmidrule(lr){2-13}
    Question type & \shortstack[c]{Input} & \shortstack[c]{Output} & \shortstack[c]{Input} & \shortstack[c]{Output} & \shortstack[c]{Input} & \shortstack[c]{Output} & \shortstack[c]{Input} & \shortstack[c]{Output} & \shortstack[c]{Input} & \shortstack[c]{Output} & \shortstack[c]{Input} & \shortstack[c]{Output} \\
    \midrule
    component count & $0.83$ & $0.54$ & $0.24$ & $0.26$ & $0.61$ & $0.27$ & $0.77$ & $0.59$ & $0.87$ & $0.89$ & $0.66$ & $0.51$ \\
    edge count & $0.75$ & $0.55$ & $0.15$ & $0.11$ & $0.26$ & $0.11$ & $0.80$ & $0.46$ & $0.97$ & $0.98$ & $0.59$ & $0.44$ \\
    has cycles & $0.76$ & $0.66$ & $0.58$ & $0.53$ & $0.49$ & $0.49$ & $0.69$ & $0.62$ & $0.97$ & $0.97$ & $0.70$ & $0.65$ \\
    is connected & $0.94$ & $0.76$ & $0.67$ & $0.64$ & $0.76$ & $0.59$ & $0.88$ & $0.77$ & $0.96$ & $0.94$ & $0.84$ & $0.74$ \\
    is tree & $0.82$ & $0.50$ & $0.57$ & $0.53$ & $0.50$ & $0.32$ & $0.80$ & $0.64$ & $0.98$ & $0.98$ & $0.73$ & $0.59$ \\
    max degree & $0.96$ & $0.57$ & $0.46$ & $0.33$ & $0.43$ & $0.28$ & $0.94$ & $0.64$ & $0.99$ & $0.96$ & $0.75$ & $0.56$ \\
    min degree & $0.96$ & $0.56$ & $0.58$ & $0.37$ & $0.49$ & $0.25$ & $0.96$ & $0.62$ & $0.98$ & $0.90$ & $0.80$ & $0.54$ \\
    node count & $0.97$ & $0.49$ & $0.63$ & $0.32$ & $0.34$ & $0.15$ & $0.96$ & $0.49$ & $1.00$ & $0.61$ & $0.78$ & $0.41$ \\
    \midrule
    full output & --- & $0.37$ & --- & $0.11$ & --- & $0.19$ & --- & $0.13$ & --- & $0.91$ & --- & $0.34$ \\
    \bottomrule
    \end{tabular}
\end{table*}

\Cref{tab:pattern_performance} summarizes performance across size patterns. The results reveal a performance drop as test graph size increases, particularly for o1-mini and o3-mini. For instance, o1-mini's accuracy declines from 88\% at 10 nodes to just 18\% at 250 nodes, suggesting a limit in working memory or attention mechanisms. In contrast, GPT 5 maintains robust performance, achieving 91\% accuracy even at 250 nodes. The scaling dataset patterns (last five rows) were only evaluated on reasoning models, as direct-answer models struggled even on the smallest graphs. 

Looking across tasks, \Cref{tab:task_by_pattern_scaling} shows a consistent decline in accuracy when increasing the test graph size. The steepest drops are observed in tasks involving degree-based deletions or colorings. For example, \texttt{removeDegree3} falls from 96\% accuracy at 10 nodes to just 25\% at 250 nodes, and similar patterns are seen for \texttt{colorDegree2,3}. Surprisingly, some tasks with more global structure, such as \texttt{colorPath} and \texttt{colorComponents}, exhibit stronger robustness, remaining near-perfect up to 100 nodes and only degrading at the largest size. Some simpler local modifications, such as \texttt{addHub} and \texttt{removeDegree1}, also show relatively stable performance compared to their higher-degree counterparts. These results highlight that scaling difficulties are not purely a function of locality: global transformations can scale well, while seemingly local ones can collapse under larger graphs, suggesting problems in how the models internalize and apply structural rules.

\paragraph{Performance in Question-Based Tasks.} Besides full graph generation, we also test model understanding with targeted questions, which isolate comprehension of graph properties from execution of transformations. Table \ref{tab:questiontype_by_model_io} shows question-based performance for selected models and question types. In general, it is easier for models to answer questions about input graphs than output graphs. The gap between high input accuracy and substantially lower output accuracy highlights that even when models can recognize graph properties, they struggle to learn the transformation or cannot apply it to new graphs reliably. See Appendix \ref{app:full results}, Table~\ref{fig:questiontype_by_model_io_full} for full results across all models and question types.

We also observe a systematic bias in reasoning models: when asked about input graphs, they often answer as if the question referred to the output graph instead. Advanced reasoning models show dramatically higher transfer rates, indicating they apply transformations even when not requested. The strength of this effect correlates positively with model capability:  
GPT 5: 55--85\%, O4-mini: 45--75\%, O3-mini: 40--70\%, Qwen3 models: $<20$\%. It is most pronounced in tasks with many coloring changes (\texttt{bipartitionCompletion} and \texttt{blueSubgraph}), and weakest in structural modifications (\texttt{addHub} and \texttt{removeDegree}).

\paragraph{Effect of Graph Encoding} Encoding choice has little average impact (performance ratio near 1.0 for most models), though preferences vary by model and task type. Most differences are minor, but Qwen3-4b is a clear outlier: for full output tasks, it achieves 31.7\% better performance with incident encoding (0.407 vs. 0.309), while for question-based tasks, it performs 24.7\% better with adjacency encoding (0.735 vs. 0.589). Overall, encoding effects appear minor on average but can be important for particular model-task combinations.

\subsection{Case Study: Failure Modes of GPT 5}
We manually analyzed the answers for GPT 5 on full output tasks. We observe that errors generally fall into the four categories:
Output parsing mismatches, Threshold misinterpretation, Concept substitution, and Encoding sensitivity.

\paragraph{(1) Output parsing mismatches.}
In some cases the model description of the solution matched the intended subgraph, but the output failed to parse correctly and was marked wrong.
\begin{quote}\small
``I can't share step-by-step reasoning, but the task is to take the induced subgraph on the colored nodes and list only edges between those colored nodes.''\\
\texttt{\string<answer\string>} G describes a graph among nodes 4, 11 \ldots\ \texttt{\string</answer\string>}
\end{quote}
\emph{Takeaway:} The semantics were correct, but formatting (e.g.\ edge listing or section tags) caused evaluation failure.

\paragraph{(2) Threshold misinterpretation.}
Tasks requiring the coloring or removal of nodes of degree $x$ (e.g. 3) led to conflicting rules: sometimes interpreted as ``degree exactly 3,'' and sometimes as ``degree at least 3.'' The examples were unambiguous, since in at least one case a node of degree four was present but left untouched, clearly indicating that the rule was ``degree exactly 3.''  
\begin{quote}\small
``\ldots removes all nodes with degree 3 or more \ldots'' \\
``\ldots remove every node with degree exactly 3 \ldots''\\
``I identify and color the node(s) with the highest degree (most connections).''
\end{quote}
\emph{Takeaway:} Even when the examples uniquely determined the threshold, the model occasionally chose a looser interpretation (``$\geq3$'') instead of the intended exact one.

\paragraph{(3) Concept substitution.}
Sometimes, the model substituted the intended property with a different but superficially related graph concept. For example:
\begin{quote}\small
``\ldots identified the articulation points (cut vertices) \ldots''
\end{quote}
\emph{Takeaway:} Instead of following the intended rule, the model sometimes defaulted to alternative structural notions that seemed plausible in one of the examples.

\paragraph{(4) Encoding sensitivity.}
Usually the two encoding formats (adjacency list vs.\ incidence list) yielded similar results. However, in some cases, particularly tasks involving distance-based reasoning and paths, the model succeeded under one encoding but failed under the other.
\begin{quote}\small
``\ldots find the unique shortest path between them. Color the path's center node red and also color any neighbors of that center \ldots'' \\
``\ldots red if and only if it is equidistant to both blue seeds; ties occur at nodes 11, 10, 9, 8, 5, 7, 4, 6, 3.''
\end{quote}
\emph{Takeaway:} The correct reasoning was present, but the execution depended on encoding format.

\section{Discussion}

Our results reveal three insights about graph reasoning in language models.  
First, we observe a persistent \emph{comprehension--execution gap}: models can parse structural properties and answer questions about them with high accuracy, but often fail to apply transformations consistently, highlighting that recognition of graph features is easier than generating coherent transformed outputs. This mirrors findings in other domains where transformers succeed at local operations but fail to compose them into globally consistent solutions~\citep{dziri2024faith}.

Secondly, scaling barriers emerge even for strong models. Mid-tier reasoning models collapse between 50--100 nodes, while GPT~5 maintains robust performance up to 250 nodes yet still struggles on seemingly local tasks such as degree-based transformations. This shows that scaling difficulty is not tied purely to locality but to how reliably models internalize and generalize transformation rules.  

Thirdly, we observe a \emph{paradox of capability}: more advanced models increasingly apply transformations even when not asked, answering questions about the input graphs as if they referred to the output. This ``transformation bias'' indicates that instruction following can degrade with capability, creating a failure mode where models over-apply their reasoning ability. This phenomenon has been observed in other contexts where highly capable models ``overthink" simple tasks \citep{wei2022chain} or exhibit ``inverse scaling" behaviors \citep{mckenzie2023inverse}.  

\subsection{Comparison with Grid-Based ARC}

GraphARC reveals complementary challenges to the original ARC benchmark:

\paragraph{Task Structure.} While ARC uses fixed small grids with visual patterns, GraphARC tests relational reasoning without spatial constraints. The absence of visual cues forces models to rely purely on structural understanding. The graph-based format allows us to automatically generate a virtually unlimited variety of instances. 

\paragraph{Interpretability.} Our decomposition into question types provides clearer failure analysis than ARC's binary success/failure. We can identify specific capabilities (input understanding, output understanding, and transformation execution) and trace failure modes to particular reasoning steps.

\paragraph{Scaling.} While ARC uses fixed small grids, our scalable approach reveals performance degradation patterns with size, exposing architectural limitations that remain hidden in fixed-size evaluations. GraphARC's ability to automatically generate large test cases allows controlled evaluation of scaling behavior and identification of failure thresholds across models. 

\paragraph{Human vs. AI Focus.} ARC was conceived as a human-intelligence challenge that is solvable for people but difficult for machines. Its grid layout provides visual cues that humans naturally exploit. In contrast, GraphARC is designed as an AI benchmark: the underlying graph structure is presented without spatial layout, since models should not depend on visualization but on reasoning over the graph's relational properties. Informal testing confirms that GraphARC tasks (for example those in \Cref{fig:example_task}) are easily solvable by humans, but this relies on having access to an appropriate graph visualization.

\section{Conclusion and Future Work}
GraphARC provides a new benchmark for studying few-shot abstract reasoning on graphs.  
By generating diverse input-output transformations across graph families and sizes, it exposes systematic limitations of current language models that remain hidden in standard reasoning benchmarks.  

Our evaluation of over 125,000 responses identifies three consistent phenomena: (1) a comprehension-execution gap where models can recognize graph properties but fail to reliably apply transformations, (2) scaling barriers that cause mid-tier models to collapse beyond 50-100 nodes, and (3) a paradoxical failure mode where more capable models over-apply transformations, even when not requested. 

We see two promising directions for future work. First, GraphARC can serve as a testbed for models explicitly designed for relational reasoning, including graph neural networks and emerging graph foundation models. Second, the benchmark invites new training strategies that move toward compositional generalization, such as curriculum design, modular reasoning approaches, or hybrid symbolic-neural methods.  

By making failure modes transparent and scalable, GraphARC aims to guide the development of systems that can reason robustly over structured data—a necessary step toward broader abstract reasoning capabilities.

\bibliographystyle{ACM-Reference-Format}
\balance
\bibliography{references}

@article{chollet2019measure,
  title={On the measure of intelligence},
  author={Chollet, Fran{\c{c}}ois},
  journal={arXiv preprint arXiv:1911.01547},
  year={2019}
}

@article{lee2024llms,
  title={Llms and the abstraction and reasoning corpus: Successes, failures, and the importance of object-based representations},
  author={Xu, Yudong and Li, Wenhao and Vaezipoor, Pashootan and Sanner, Scott and Khalil, Elias B},
  journal={arXiv preprint arXiv:2305.18354},
  year={2023}
}

@inproceedigs{fatemi2023talk,
  title={Talk like a Graph: Encoding Graphs for Large Language Models},
  author={Bahare Fatemi and Jonathan Halcrow and Bryan Perozzi},
  booktitle={International Conference on Learning Representations (ICLR)},
  year={2024}
}

@article{wang2023can,
  title={Can language models solve graph problems in natural language?},
  author={Wang, Heng and Feng, Shangbin and He, Tianxing and Tan, Zhaoxuan and Han, Xiaochuang and Tsvetkov, Yulia},
  journal={Advances in Neural Information Processing Systems},
  volume={36},
  pages={30840--30861},
  year={2023}
}

@article{liu2023towards,
  title={Towards graph foundation models: A survey and beyond},
  author={Liu, Jiawei and Yang, Cheng and Lu, Zhiyuan and Chen, Junze and Li, Yibo and Zhang, Mengmei and Bai, Ting and Fang, Yuan and Sun, Lichao and Yu, Philip S and others},
  journal={arXiv preprint arXiv:2310.11829},
  year={2023}
}

@misc{arcprize2024,
      title={OpenAI O3 Breakthrough High Score on ARC-AGI-PUB},
      author={{ARC Prize Foundation}},
      year={2024},
      howpublished={\url{https://arcprize.org/blog/oai-o3-pub-breakthrough}},
      note={Accessed: 2025-01-01}
}

@article{jin2024large,
author = {Jin, Bowen and Liu, Gang and Han, Chi and Jiang, Meng and Ji, Heng and Han, Jiawei},
title = {Large Language Models on Graphs: A Comprehensive Survey},
year = {2024},
issue_date = {Dec. 2024},
publisher = {IEEE Educational Activities Department},
address = {USA},
volume = {36},
number = {12},
issn = {1041-4347},
url = {https://doi.org/10.1109/TKDE.2024.3469578},
doi = {10.1109/TKDE.2024.3469578},
journal = {IEEE Trans. on Knowl. and Data Eng.},
month = dec,
pages = {8622–8642},
numpages = {21}
}

@article{dai2024large,
  title={How do large language models understand graph patterns? a benchmark for graph pattern comprehension},
  author={Dai, Xinnan and Qu, Haohao and Shen, Yifen and Zhang, Bohang and Wen, Qihao and Fan, Wenqi and Li, Dongsheng and Tang, Jiliang and Shan, Caihua},
  journal={arXiv preprint arXiv:2410.05298},
  year={2024}
}

@article{mckenzie2023inverse,
  title={Inverse scaling: When bigger isn't better},
  author={McKenzie, Ian R and Lyzhov, Alexander and Pieler, Michael and Parrish, Alicia and Mueller, Aaron and Prabhu, Ameya and McLean, Euan and Kirtland, Aaron and Ross, Alexis and Liu, Alisa and others},
  journal={arXiv preprint arXiv:2306.09479},
  year={2023}
}

@inproceedings{wei2022chain,
author = {Wei, Jason and Wang, Xuezhi and Schuurmans, Dale and Bosma, Maarten and Ichter, Brian and Xia, Fei and Chi, Ed H. and Le, Quoc V. and Zhou, Denny},
title = {Chain-of-thought prompting elicits reasoning in large language models},
year = {2022},
isbn = {9781713871088},
publisher = {Curran Associates Inc.},
address = {Red Hook, NY, USA},
booktitle = {Proceedings of the 36th International Conference on Neural Information Processing Systems},
articleno = {1800},
numpages = {14},
location = {New Orleans, LA, USA},
series = {NIPS '22}
}

@inproceedings{dziri2024faith,
author = {Dziri, Nouha and Lu, Ximing and Sclar, Melanie and Li, Xiang Lorraine and Jiang, Liwei and Lin, Bill Yuchen and West, Peter and Bhagavatula, Chandra and Le Bras, Ronan and Hwang, Jena D. and Sanyal, Soumya and Welleck, Sean and Ren, Xiang and Ettinger, Allyson and Harchaoui, Zaid and Choi, Yejin},
title = {Faith and fate: limits of transformers on compositionality},
year = {2023},
publisher = {Curran Associates Inc.},
address = {Red Hook, NY, USA},
booktitle = {Proceedings of the 37th International Conference on Neural Information Processing Systems},
articleno = {3081},
numpages = {40},
location = {New Orleans, LA, USA},
series = {NIPS '23}
}

@inproceedings{assouel2022object,
  title={Object-centric compositional imagination for visual abstract reasoning},
  author={Assouel, Rim and Rodriguez, Pau and Taslakian, Perouz and Vazquez, David and Bengio, Yoshua},
  booktitle={ICLR2022 Workshop on the Elements of Reasoning: Objects, Structure and Causality},
  year={2022}
}

@article{HALFORD2010497,
  title={Relational knowledge: The foundation of higher cognition},
  author={Halford, Graeme S and Wilson, William H and Phillips, Steven},
  journal={Trends in cognitive sciences},
  volume={14},
  number={11},
  pages={497--505},
  year={2010},
  publisher={Elsevier}
}

@article{holyoak201213,
  title={13 Analogy and Relational Reasoning},
  author={Holyoak, Keith J},
  journal={The Oxford handbook of thinking and reasoning},
  pages={234},
  year={2012},
  publisher={Oxford University Press}
}

@article{smith1992case,
  title={The case for rules in reasoning},
  author={Smith, Edward E and Langston, Christopher and Nisbett, Richard E},
  journal={Cognitive science},
  volume={16},
  number={1},
  pages={1--40},
  year={1992},
  publisher={Wiley Online Library}
}

@article{hummel2003symbolic,
  title={A symbolic-connectionist theory of relational inference and generalization.},
  author={Hummel, John E and Holyoak, Keith J},
  journal={Psychological review},
  volume={110},
  number={2},
  pages={220},
  year={2003},
  publisher={American Psychological Association}
}

@article{pinker1998words,
  title={Words and rules},
  author={Pinker, Steven},
  journal={Lingua},
  volume={106},
  number={1-4},
  pages={219--242},
  year={1998},
  publisher={Elsevier}
}

@inproceedings{kipf2016semi,
  author       = {Thomas N. Kipf and
                  Max Welling},
  title        = {Semi-Supervised Classification with Graph Convolutional Networks},
  booktitle    = {5th International Conference on Learning Representations, {ICLR} 2017,
                  Toulon, France, April 24-26, 2017, Conference Track Proceedings},
  publisher    = {OpenReview.net},
  year         = {2017},
  url          = {https://openreview.net/forum?id=SJU4ayYgl},
  bibsource    = {dblp computer science bibliography, https://dblp.org}
}

@article{ying2018hierarchical,
  title={Hierarchical graph representation learning with differentiable pooling},
  author={Ying, Zhitao and You, Jiaxuan and Morris, Christopher and Ren, Xiang and Hamilton, Will and Leskovec, Jure},
  journal={Advances in neural information processing systems},
  volume={31},
  year={2018}
}

@article{zhang2018link,
  title={Link prediction based on graph neural networks},
  author={Zhang, Muhan and Chen, Yixin},
  journal={Advances in neural information processing systems},
  volume={31},
  year={2018}
}

@inproceedings{simonovsky2018graphvae,
  title={Graphvae: Towards generation of small graphs using variational autoencoders},
  author={Simonovsky, Martin and Komodakis, Nikos},
  booktitle={International conference on artificial neural networks},
  pages={412--422},
  year={2018},
  organization={Springer}
}

@article{liu2025graph,
  title={Graph foundation models: Concepts, opportunities and challenges},
  author={Liu, Jiawei and Yang, Cheng and Lu, Zhiyuan and Chen, Junze and Li, Yibo and Zhang, Mengmei and Bai, Ting and Fang, Yuan and Sun, Lichao and Yu, Philip S and others},
  journal={IEEE Transactions on Pattern Analysis and Machine Intelligence},
  year={2025},
  publisher={IEEE}
}

@article{wang2025graph,
  title={Graph Foundation Models: A Comprehensive Survey},
  author={Wang, Zehong and Liu, Zheyuan and Ma, Tianyi and Li, Jiazheng and Zhang, Zheyuan and Fu, Xingbo and Li, Yiyang and Yuan, Zhengqing and Song, Wei and Ma, Yijun and others},
  journal={arXiv preprint arXiv:2505.15116},
  year={2025}
}

@article{lake2017building,
  title={Building machines that learn and think like people},
  author={Lake, Brenden M and Ullman, Tomer D and Tenenbaum, Joshua B and Gershman, Samuel J},
  journal={Behavioral and brain sciences},
  volume={40},
  pages={e253},
  year={2017},
  publisher={Cambridge University Press}
}

@article{chollet2025arc,
  title={Arc-agi-2: A new challenge for frontier ai reasoning systems},
  author={Chollet, Francois and Knoop, Mike and Kamradt, Gregory and Landers, Bryan and Pinkard, Henry},
  journal={arXiv preprint arXiv:2505.11831},
  year={2025}
}

@article{zhao2023gimlet,
  title={Gimlet: A unified graph-text model for instruction-based molecule zero-shot learning},
  author={Zhao, Haiteng and Liu, Shengchao and Chang, Ma and Xu, Hannan and Fu, Jie and Deng, Zhihong and Kong, Lingpeng and Liu, Qi},
  journal={Advances in neural information processing systems},
  volume={36},
  pages={5850--5887},
  year={2023}
}

@misc{arcprize2024progress,
  title        = {2024 Progress on ARC-AGI-Pub},
  author       = {{ARC Prize Foundation}},
  year         = {2024},
  howpublished = {\url{https://arcprize.org/blog/2024-progress-arc-agi-pub}},
  note         = {Accessed: 2025-09-01}
}

@article{galkin2024ultraquery,
  title={A foundation model for zero-shot logical query reasoning},
  author={Galkin, Michael and Zhou, Jincheng and Ribeiro, Bruno and Tang, Jian and Zhu, Zhaocheng},
  journal={Advances in Neural Information Processing Systems},
  volume={37},
  pages={54137--54160},
  year={2024}
}

@article{mole,
  title={MolE: a foundation model for molecular graphs using disentangled attention},
  author={M{\'e}ndez-Lucio, Oscar and Nicolaou, Christos A and Earnshaw, Berton},
  journal={Nature Communications},
  volume={15},
  number={1},
  pages={9431},
  year={2024},
  publisher={Nature Publishing Group UK London}
}

@article{sanford2024understanding,
  title={Understanding transformer reasoning capabilities via graph algorithms},
  author={Sanford, Clayton and Fatemi, Bahare and Hall, Ethan and Tsitsulin, Anton and Kazemi, Mehran and Halcrow, Jonathan and Perozzi, Bryan and Mirrokni, Vahab},
  journal={Advances in Neural Information Processing Systems},
  volume={37},
  pages={78320--78370},
  year={2024}
}

@article{mirtaheri2025let,
  title={Let Me Think! A Long Chain-of-Thought Can Be Worth Exponentially Many Short Ones},
  author={Mirtaheri, Parsa and Edelman, Ezra and Jelassi, Samy and Malach, Eran and Boix-Adsera, Enric},
  journal={arXiv preprint arXiv:2505.21825},
  year={2025}
}

@inproceedings{snell2025scaling,
  title={Scaling LLM test-time compute optimally can be more effective than scaling parameters for reasoning},
  author={Snell, Charlie Victor and Lee, Jaehoon and Xu, Kelvin and Kumar, Aviral},
  booktitle={The Thirteenth International Conference on Learning Representations},
  year={2025}
}

\appendix

\section{Instance Generation Framework}
\label{app:generation_validation}

\subsection{Graph Generators}
\label{app:graph_generators}

We employ multiple graph generators to ensure comprehensive coverage of structural patterns:

\begin{itemize}
\item \textbf{Random}: Erdős-Rényi $G(n,p)$ model with edge probability $p=0.3$
\item \textbf{Connected}: Watts-Strogatz small-world graphs (guaranteed connected)
\item \textbf{Trees}: Generated by extracting a BFS spanning tree from a connected Watts–Strogatz small-world graph 
\item \textbf{Star}: Center node connected to $n-1$ leaves
\item \textbf{Bipartite}: Two-partition graphs with random inter-partition edges
\item \textbf{Multi-Component}: Exactly 2 disconnected components
\end{itemize}

\subsection{Size Patterns}
\label{app: patterns}

Size patterns control how few-shot learning occurs by varying \textbf{both the number of examples and their size}. We design two pattern groups to isolate different learning challenges:

\textit{Main Dataset Patterns} (varying number of examples):
\begin{itemize}
\item \texttt{scale\_up\_3} [5, 10, 15]: Test on 15-node graph after seeing a 5-node and a 10-node example
\item \texttt{scale\_up\_4} [5, 10, 15, 15]: Test on 15-node graph after seeing a 5-node, a 10-node, and a 15-node example
\end{itemize}

\textit{Scaling Dataset Patterns} (testing final graph size):
\begin{itemize}
\item \texttt{cap10\_3} [10, 10, 10]: Baseline with all 10-node graphs
\item \texttt{cap25\_3} [10, 10, 25]: Test on 25-node graph after seeing two 10-node examples
\item \texttt{cap50\_3} [10, 10, 50]: Test on 50-node graph after 2 10-node examples
\item \texttt{cap100\_3} [10, 10, 100]: Test on 100-node graph after seeing two 10-node examples
\item \texttt{cap250\_3} [10, 10, 250]: Test on 250-node graph after seeing two 10-node examples
\end{itemize}

\subsection{Property validation system}
\label{app:validation}

To ensure task validity and meaningful evaluation, we implement a comprehensive property validation system. Each task declares required graph properties (e.g., connectivity, specific degree distributions), and our generation pipeline validates that produced graphs satisfy these constraints.

The validation system uses a three-state property framework: 
\begin{itemize}
\item \textbf{TRUE}: Property is guaranteed by the generator
\item \textbf{FALSE}: Property is incompatible with the generator  
\item \textbf{MAYBE}: Property may or may not hold depending on random generation
\end{itemize}

This framework enables efficient validation by skipping guaranteed properties and immediately rejecting incompatible generator-transformation combinations. Additionally, we verify that each transformation produces well-defined transformations on the example graphs, ensuring that the learning problem is neither trivial nor ill-posed.

\section{Experimental Design for LLM Evaluation}
\label{app: encoding}

This section outlines how we evaluate GraphARC on language models. We test the models separately on the full output and question-based pathways. We also test the effect of graph encoding schemes and system prompts on model performance. 

\subsection{Graph Encoding Schemes}

Graph structures must be converted to text for language model processing. We implement two encoding schemes that emphasize different structural properties:

\textbf{Adjacency List Format}: This encoding lists all edges as node pairs. It may facilitate reasoning about paths and global properties. An example encoding for a path graph is shown below:

\begin{quote}
    In an undirected graph, (i,j) means that node i and node j are connected with an undirected edge.\\
    G describes a graph among nodes 0, 1, 2, 3, 4.\\
    The edges in G are: (0,1) (1,2) (2,3) (3,4).\\
    The following nodes are colored: 1, 2, 3.
\end{quote}
    
\textbf{Incidence List Format}: This encoding lists all nodes in the immediate neighborhood for each node. Degree and local transformation based tasks may benefit from this format.

\begin{quote}
G describes a graph among nodes 0, 1, 2, 3, 4.\\
In this graph:\\
Node 0 is connected to nodes 1.\\
Node 1 is connected to nodes 0, 2.\\
Node 2 is connected to nodes 1, 3.\\
Node 3 is connected to nodes 2, 4.\\
Node 4 is connected to nodes 3.\\
The following nodes are colored: 1, 2, 3.
\end{quote}

\subsection{System Prompt Variation}\label{app: system_prompts_patterns}

We experiment with 3 different system prompts that frame the task from different professional perspectives, as well as a baseline with no system prompt. \textit{Analyst} emphasizes systematic analysis, \textit{Programmer} invokes algorithmic thinking, and \textit{Teacher} encourages explanation. The full prompts are shown below:

\begin{table}[ht]
    \label{table-system-prompts}
    \centering
    \begin{tabular}{p{0.18\linewidth} p{0.74\linewidth}}
    \hline
    \textbf{Role} & \textbf{Prompt} \\
    \hline
    Analyst &
    "You are a graph analyst. Study the following graph examples carefully and answer the question that follows." \\
    Programmer &
    "You are a graph algorithm developer. Analyze the example graphs and their patterns, then answer the question about the given input." \\
    Teacher &
    "You are a mathematics teacher. Examine these graph examples to understand any patterns, then answer the question clearly and methodically." \\
    None & No system prompt provided \\
    \hline
    \end{tabular}
    \end{table}

System prompts show minimal and inconsistent effects across models. Most models vary less than 5\% between prompts, with no prompt consistently outperforming others. Qwen3-4b again shows highest sensitivity (up to 23.8\% variation), but the pattern is model-specific rather than systematic. Notably, the ``none" baseline equals or exceeds role-based prompts several times, suggesting that explicit role assignment may interfere with models' natural reasoning strategies for abstract tasks.

\section{Full results}\label{app:full results}

\begin{table*}[t]
    \centering
    \caption{Accuracy on full-output tasks across models. Values are mean accuracies with 95\% confidence intervals (computed using \texttt{scipy.stats.bootstrap} with percentile method).}
    \label{tab:task_performance_with_ci}
    % First table
    \begin{subtable}{\textwidth}
        \centering
        \small
    \setlength{\tabcolsep}{1mm}  %%
    \begin{tabular}{lcccccccc}
        \toprule
        Task & \shortstack[c]{LLaMA3\\8b} & \shortstack[c]{LLaMA3.1\\8b} & \shortstack[c]{DeepSeek-R1\\LLaMA-8b} & \shortstack[c]{GPT\\4.1-nano} & \shortstack[c]{o1\\mini} & \shortstack[c]{o3\\mini} & \shortstack[c]{o4\\mini} & \shortstack[c]{GPT\\5} \\
        \midrule
        removeDegree3 & $0.00 \pm 0.00$ & $0.00 \pm 0.00$ & $0.00 \pm 0.00$ & $0.00 \pm 0.00$ & $0.00 \pm 0.00$ & $0.25 \pm 0.25$ & $0.50 \pm 0.29$ & $0.75 \pm 0.23$ \\
        colorDegree3 & $0.00 \pm 0.00$ & $0.00 \pm 0.00$ & $0.10 \pm 0.09$ & $0.00 \pm 0.00$ & $0.00 \pm 0.00$ & $0.67 \pm 0.27$ & $0.42 \pm 0.25$ & $0.58 \pm 0.25$ \\
        colorEquidistant & $0.00 \pm 0.00$ & $0.00 \pm 0.00$ & $0.00 \pm 0.00$ & $0.00 \pm 0.00$ & $0.38 \pm 0.31$ & $0.50 \pm 0.38$ & $0.75 \pm 0.31$ & $0.75 \pm 0.25$ \\
        colorMaxDegree & $0.03 \pm 0.05$ & $0.25 \pm 0.14$ & $0.04 \pm 0.05$ & $0.25 \pm 0.12$ & $0.25 \pm 0.25$ & $0.33 \pm 0.25$ & $0.58 \pm 0.25$ & $0.67 \pm 0.27$ \\
        colorDistanceAtLeast2 & $0.00 \pm 0.00$ & $0.00 \pm 0.00$ & $0.00 \pm 0.00$ & $0.00 \pm 0.00$ & $0.08 \pm 0.12$ & $0.83 \pm 0.21$ & $0.92 \pm 0.12$ & $0.92 \pm 0.12$ \\
        removeDegree2 & $0.00 \pm 0.00$ & $0.00 \pm 0.00$ & $0.00 \pm 0.00$ & $0.00 \pm 0.00$ & $0.42 \pm 0.29$ & $0.75 \pm 0.25$ & $0.92 \pm 0.12$ & $1.00 \pm 0.00$ \\
        edgeToNode & $0.00 \pm 0.00$ & $0.00 \pm 0.00$ & $0.00 \pm 0.00$ & $0.00 \pm 0.00$ & $0.50 \pm 0.25$ & $0.88 \pm 0.16$ & $0.94 \pm 0.09$ & $1.00 \pm 0.00$ \\
        colorMinDegree & $0.00 \pm 0.00$ & $0.02 \pm 0.03$ & $0.29 \pm 0.12$ & $0.12 \pm 0.09$ & $0.67 \pm 0.25$ & $0.75 \pm 0.25$ & $0.67 \pm 0.25$ & $0.83 \pm 0.21$ \\
        colorDegree2 & $0.00 \pm 0.00$ & $0.00 \pm 0.00$ & $0.06 \pm 0.07$ & $0.04 \pm 0.05$ & $0.42 \pm 0.25$ & $0.92 \pm 0.12$ & $1.00 \pm 0.00$ & $1.00 \pm 0.00$ \\
        mergeAtBlue & $0.00 \pm 0.00$ & $0.00 \pm 0.00$ & $0.00 \pm 0.00$ & $0.00 \pm 0.00$ & $0.75 \pm 0.38$ & $1.00 \pm 0.00$ & $1.00 \pm 0.00$ & $1.00 \pm 0.00$ \\
        removeSameColorEdges & $0.00 \pm 0.00$ & $0.02 \pm 0.04$ & $0.21 \pm 0.10$ & $0.02 \pm 0.03$ & $0.75 \pm 0.23$ & $0.75 \pm 0.25$ & $1.00 \pm 0.00$ & $1.00 \pm 0.00$ \\
        bipartitionCompletion & $0.00 \pm 0.00$ & $0.00 \pm 0.00$ & $0.00 \pm 0.00$ & $0.00 \pm 0.00$ & $0.75 \pm 0.38$ & $1.00 \pm 0.00$ & $1.00 \pm 0.00$ & $1.00 \pm 0.00$ \\
        blueSubgraph & $0.00 \pm 0.00$ & $0.03 \pm 0.04$ & $0.56 \pm 0.12$ & $0.17 \pm 0.09$ & $0.75 \pm 0.20$ & $0.75 \pm 0.20$ & $0.75 \pm 0.19$ & $0.75 \pm 0.19$ \\
        complementGraph & $0.08 \pm 0.07$ & $0.17 \pm 0.09$ & $0.16 \pm 0.09$ & $0.05 \pm 0.05$ & $0.69 \pm 0.22$ & $1.00 \pm 0.00$ & $0.94 \pm 0.12$ & $1.00 \pm 0.00$ \\
        colorInternal & $0.23 \pm 0.13$ & $0.23 \pm 0.09$ & $0.20 \pm 0.10$ & $0.25 \pm 0.09$ & $0.62 \pm 0.22$ & $1.00 \pm 0.00$ & $0.88 \pm 0.16$ & $1.00 \pm 0.00$ \\
        colorComponents & $0.00 \pm 0.00$ & $0.06 \pm 0.09$ & $0.44 \pm 0.25$ & $0.19 \pm 0.19$ & $1.00 \pm 0.00$ & $0.75 \pm 0.38$ & $1.00 \pm 0.00$ & $1.00 \pm 0.00$ \\
        colorNeighbors & $0.02 \pm 0.04$ & $0.19 \pm 0.09$ & $0.14 \pm 0.08$ & $0.44 \pm 0.12$ & $0.94 \pm 0.09$ & $0.94 \pm 0.09$ & $1.00 \pm 0.00$ & $0.88 \pm 0.16$ \\
        colorPath & $0.15 \pm 0.15$ & $0.16 \pm 0.12$ & $0.09 \pm 0.09$ & $0.31 \pm 0.16$ & $0.88 \pm 0.19$ & $1.00 \pm 0.00$ & $1.00 \pm 0.00$ & $1.00 \pm 0.00$ \\
        removeDegree1 & $0.13 \pm 0.10$ & $0.19 \pm 0.09$ & $0.34 \pm 0.10$ & $0.19 \pm 0.10$ & $0.75 \pm 0.22$ & $1.00 \pm 0.00$ & $1.00 \pm 0.00$ & $1.00 \pm 0.00$ \\
        colorDegree1 & $0.09 \pm 0.09$ & $0.25 \pm 0.11$ & $0.45 \pm 0.12$ & $0.27 \pm 0.11$ & $0.88 \pm 0.16$ & $1.00 \pm 0.00$ & $1.00 \pm 0.00$ & $1.00 \pm 0.00$ \\
        addHub & $0.12 \pm 0.09$ & $0.38 \pm 0.12$ & $0.50 \pm 0.12$ & $0.20 \pm 0.09$ & $0.75 \pm 0.20$ & $1.00 \pm 0.00$ & $1.00 \pm 0.00$ & $1.00 \pm 0.00$ \\
        \midrule
        average & $0.04$ & $0.09$ & $0.17$ & $0.12$ & $0.58$ & $0.81$ & $0.87$ & $0.91$ \\
        \bottomrule
        \end{tabular}
    \end{subtable}

    \vspace{1em}

    \begin{subtable}{\textwidth}
        \centering
        \small
    \setlength{\tabcolsep}{1mm}  %%
    \begin{tabular}{lccccccc}
        \toprule
        Task & \shortstack[c]{Qwen3\\1.7b} & \shortstack[c]{Qwen3\\4b} & \shortstack[c]{Qwen3\\8b} & \shortstack[c]{Qwen3\\14b} & \shortstack[c]{Qwen3\\32b} & \shortstack[c]{Mistral\\7b-v0.2} & \shortstack[c]{OLMo-2\\7b} \\
        \midrule
        bipartitionCompletion & $0.00 \pm 0.00$ & $0.00 \pm 0.00$ & $0.00 \pm 0.00$ & $0.00 \pm 0.00$ & $0.00 \pm 0.00$ & $0.00 \pm 0.00$ & $0.00 \pm 0.00$ \\
        removeDegree2 & $0.00 \pm 0.00$ & $0.00 \pm 0.00$ & $0.00 \pm 0.00$ & $0.00 \pm 0.00$ & $0.00 \pm 0.00$ & $0.00 \pm 0.00$ & $0.00 \pm 0.00$ \\
        colorDistanceAtLeast2 & $0.00 \pm 0.00$ & $0.00 \pm 0.00$ & $0.00 \pm 0.00$ & $0.00 \pm 0.00$ & $0.00 \pm 0.00$ & $0.00 \pm 0.00$ & $0.00 \pm 0.00$ \\
        removeDegree3 & $0.00 \pm 0.00$ & $0.00 \pm 0.00$ & $0.00 \pm 0.00$ & $0.02 \pm 0.03$ & $0.02 \pm 0.03$ & $0.00 \pm 0.00$ & $0.00 \pm 0.00$ \\
        mergeAtBlue & $0.00 \pm 0.00$ & $0.00 \pm 0.00$ & $0.00 \pm 0.00$ & $0.00 \pm 0.00$ & $0.12 \pm 0.16$ & $0.00 \pm 0.00$ & $0.00 \pm 0.00$ \\
        colorEquidistant & $0.03 \pm 0.05$ & $0.00 \pm 0.00$ & $0.00 \pm 0.00$ & $0.00 \pm 0.00$ & $0.16 \pm 0.12$ & $0.00 \pm 0.00$ & $0.03 \pm 0.05$ \\
        colorDegree3 & $0.02 \pm 0.03$ & $0.10 \pm 0.08$ & $0.10 \pm 0.09$ & $0.29 \pm 0.12$ & $0.06 \pm 0.06$ & $0.00 \pm 0.00$ & $0.00 \pm 0.00$ \\
        complementGraph & $0.02 \pm 0.02$ & $0.11 \pm 0.08$ & $0.11 \pm 0.07$ & $0.19 \pm 0.09$ & $0.17 \pm 0.09$ & $0.00 \pm 0.00$ & $0.00 \pm 0.00$ \\
        colorDegree2 & $0.00 \pm 0.00$ & $0.04 \pm 0.05$ & $0.02 \pm 0.03$ & $0.42 \pm 0.15$ & $0.15 \pm 0.09$ & $0.00 \pm 0.00$ & $0.00 \pm 0.00$ \\
        edgeToNode & $0.00 \pm 0.00$ & $0.02 \pm 0.02$ & $0.08 \pm 0.06$ & $0.25 \pm 0.11$ & $0.33 \pm 0.12$ & $0.00 \pm 0.00$ & $0.00 \pm 0.00$ \\
        removeSameColorEdges & $0.00 \pm 0.00$ & $0.17 \pm 0.09$ & $0.23 \pm 0.12$ & $0.46 \pm 0.14$ & $0.48 \pm 0.14$ & $0.00 \pm 0.00$ & $0.00 \pm 0.00$ \\
        removeDegree1 & $0.16 \pm 0.09$ & $0.30 \pm 0.10$ & $0.28 \pm 0.11$ & $0.41 \pm 0.12$ & $0.36 \pm 0.11$ & $0.10 \pm 0.08$ & $0.03 \pm 0.04$ \\
        colorInternal & $0.20 \pm 0.09$ & $0.34 \pm 0.12$ & $0.38 \pm 0.11$ & $0.44 \pm 0.12$ & $0.27 \pm 0.10$ & $0.04 \pm 0.05$ & $0.05 \pm 0.05$ \\
        colorMinDegree & $0.08 \pm 0.07$ & $0.27 \pm 0.13$ & $0.40 \pm 0.14$ & $0.58 \pm 0.14$ & $0.60 \pm 0.12$ & $0.00 \pm 0.00$ & $0.00 \pm 0.00$ \\
        colorMaxDegree & $0.10 \pm 0.08$ & $0.48 \pm 0.12$ & $0.46 \pm 0.14$ & $0.52 \pm 0.14$ & $0.38 \pm 0.13$ & $0.05 \pm 0.06$ & $0.00 \pm 0.00$ \\
        colorPath & $0.25 \pm 0.16$ & $0.44 \pm 0.17$ & $0.47 \pm 0.15$ & $0.41 \pm 0.17$ & $0.50 \pm 0.18$ & $0.03 \pm 0.05$ & $0.00 \pm 0.00$ \\
        colorComponents & $0.00 \pm 0.00$ & $0.31 \pm 0.22$ & $0.62 \pm 0.25$ & $0.62 \pm 0.22$ & $0.50 \pm 0.25$ & $0.07 \pm 0.10$ & $0.00 \pm 0.00$ \\
        colorDegree1 & $0.12 \pm 0.08$ & $0.55 \pm 0.12$ & $0.62 \pm 0.12$ & $0.88 \pm 0.09$ & $0.72 \pm 0.11$ & $0.04 \pm 0.04$ & $0.05 \pm 0.05$ \\
        blueSubgraph & $0.22 \pm 0.10$ & $0.67 \pm 0.12$ & $0.70 \pm 0.12$ & $0.75 \pm 0.11$ & $0.73 \pm 0.11$ & $0.00 \pm 0.00$ & $0.03 \pm 0.04$ \\
        colorNeighbors & $0.05 \pm 0.05$ & $0.73 \pm 0.11$ & $0.56 \pm 0.12$ & $0.88 \pm 0.09$ & $0.88 \pm 0.08$ & $0.04 \pm 0.05$ & $0.03 \pm 0.04$ \\
        addHub & $0.25 \pm 0.11$ & $0.83 \pm 0.09$ & $0.77 \pm 0.10$ & $0.66 \pm 0.12$ & $0.66 \pm 0.11$ & $0.06 \pm 0.07$ & $0.00 \pm 0.00$ \\
        \midrule
        average & $0.07$ & $0.26$ & $0.28$ & $0.37$ & $0.34$ & $0.02$ & $0.01$ \\
        \bottomrule
        \end{tabular}
    \end{subtable}

\end{table*}

\begin{table*}[t]
    \centering
    \caption{Performance on question-based tasks on the input and output graphs across models. Values are mean accuracies.}
    \label{fig:questiontype_by_model_io_full}
    % First table
    \begin{subtable}{\textwidth}
        \centering
        \small
    \setlength{\tabcolsep}{1mm}  %%
        \begin{tabular}{lcccccccccc}
        \toprule
        & \multicolumn{2}{c}{\shortstack[c]{Qwen3\\1.7b}} & \multicolumn{2}{c}{\shortstack[c]{Qwen3\\4b}} & \multicolumn{2}{c}{\shortstack[c]{Qwen3\\8b}} & \multicolumn{2}{c}{\shortstack[c]{Qwen3\\14b}} & \multicolumn{2}{c}{\shortstack[c]{Qwen3\\32b}} \\
        \cmidrule(lr){2-11}
        Question type & \shortstack[c]{Input} & \shortstack[c]{Output} & \shortstack[c]{Input} & \shortstack[c]{Output} & \shortstack[c]{Input} & \shortstack[c]{Output} & \shortstack[c]{Input} & \shortstack[c]{Output} & \shortstack[c]{Input} & \shortstack[c]{Output} \\
        \midrule
        component count & $0.71$ & $0.59$ & $0.93$ & $0.55$ & $0.94$ & $0.62$ & $0.96$ & $0.56$ & $0.83$ & $0.54$ \\
        edge count & $0.33$ & $0.27$ & $0.82$ & $0.54$ & $0.91$ & $0.61$ & $0.98$ & $0.67$ & $0.75$ & $0.55$ \\
        has cycles & $0.63$ & $0.53$ & $0.65$ & $0.57$ & $0.88$ & $0.80$ & $0.91$ & $0.85$ & $0.76$ & $0.66$ \\
        is connected & $0.81$ & $0.61$ & $0.80$ & $0.68$ & $0.99$ & $0.86$ & $1.00$ & $0.90$ & $0.94$ & $0.76$ \\
        is tree & $0.65$ & $0.58$ & $0.67$ & $0.45$ & $0.93$ & $0.67$ & $0.98$ & $0.70$ & $0.82$ & $0.50$ \\
        max degree & $0.62$ & $0.46$ & $0.88$ & $0.50$ & $0.96$ & $0.62$ & $0.98$ & $0.62$ & $0.96$ & $0.57$ \\
        min degree & $0.78$ & $0.51$ & $0.93$ & $0.54$ & $0.98$ & $0.59$ & $0.98$ & $0.63$ & $0.96$ & $0.56$ \\
        node count & $0.62$ & $0.39$ & $0.92$ & $0.30$ & $0.99$ & $0.62$ & $0.99$ & $0.39$ & $0.97$ & $0.49$ \\
        \midrule
        full output & --- & $0.08$ & --- & $0.29$ & --- & $0.30$ & --- & $0.41$ & --- & $0.37$ \\
        \bottomrule
        \end{tabular}
    \end{subtable}

    \vspace{1em}
    
    \begin{subtable}{\textwidth}
        \centering
        \small
    \setlength{\tabcolsep}{1mm}  %%
        \begin{tabular}{lcccccccccc}
        \toprule
        & \multicolumn{2}{c}{\shortstack[c]{GPT\\4.1-nano}} & \multicolumn{2}{c}{\shortstack[c]{o1\\mini}} & \multicolumn{2}{c}{\shortstack[c]{o3\\mini}} & \multicolumn{2}{c}{\shortstack[c]{o4\\mini}} & \multicolumn{2}{c}{\shortstack[c]{GPT\\5}} \\
        \cmidrule(lr){2-11}
        Question type & \shortstack[c]{Input} & \shortstack[c]{Output} & \shortstack[c]{Input} & \shortstack[c]{Output} & \shortstack[c]{Input} & \shortstack[c]{Output} & \shortstack[c]{Input} & \shortstack[c]{Output} & \shortstack[c]{Input} & \shortstack[c]{Output} \\
        \midrule
        component count & $0.77$ & $0.59$ & $0.94$ & $0.65$ & $0.95$ & $0.79$ & $0.91$ & $0.89$ & $0.87$ & $0.89$ \\
        edge count & $0.80$ & $0.46$ & $0.92$ & $0.76$ & $1.00$ & $0.94$ & $1.00$ & $0.98$ & $0.97$ & $0.98$ \\
        has cycles & $0.69$ & $0.62$ & $0.97$ & $0.90$ & $0.99$ & $0.95$ & $0.96$ & $0.94$ & $0.97$ & $0.97$ \\
        is connected & $0.88$ & $0.77$ & $1.00$ & $0.92$ & $1.00$ & $0.98$ & $0.98$ & $0.97$ & $0.96$ & $0.94$ \\
        is tree & $0.80$ & $0.64$ & $0.94$ & $0.93$ & $0.98$ & $0.99$ & $0.99$ & $0.98$ & $0.98$ & $0.98$ \\
        max degree & $0.94$ & $0.64$ & $1.00$ & $0.85$ & $1.00$ & $0.92$ & $1.00$ & $0.97$ & $0.99$ & $0.96$ \\
        min degree & $0.96$ & $0.62$ & $0.99$ & $0.79$ & $1.00$ & $0.93$ & $1.00$ & $0.95$ & $0.98$ & $0.90$ \\
        node count & $0.96$ & $0.49$ & $0.99$ & $0.63$ & $1.00$ & $0.83$ & $1.00$ & $0.69$ & $1.00$ & $0.61$ \\
        \midrule
        full output & --- & $0.13$ & --- & $0.58$ & --- & $0.82$ & --- & $0.87$ & --- & $0.91$ \\
        \bottomrule
        \end{tabular}
    \end{subtable}

    \vspace{1em}

    \begin{subtable}{\textwidth}
        \centering
        \small
    \setlength{\tabcolsep}{1mm}  %%
        \begin{tabular}{lcccccccccc}
        \toprule
        & \multicolumn{2}{c}{\shortstack[c]{LLaMA3\\8b}} & \multicolumn{2}{c}{\shortstack[c]{LLaMA3.1\\8b}} & \multicolumn{2}{c}{\shortstack[c]{DeepSeek-R1\\LLaMA-8b}} & \multicolumn{2}{c}{\shortstack[c]{Mistral\\7b-v0.2}} & \multicolumn{2}{c}{\shortstack[c]{OLMo-2\\7b}} \\
        \cmidrule(lr){2-11}
        Question type & \shortstack[c]{Input} & \shortstack[c]{Output} & \shortstack[c]{Input} & \shortstack[c]{Output} & \shortstack[c]{Input} & \shortstack[c]{Output} & \shortstack[c]{Input} & \shortstack[c]{Output} & \shortstack[c]{Input} & \shortstack[c]{Output} \\
        \midrule
        component count & $0.12$ & $0.16$ & $0.24$ & $0.26$ & $0.61$ & $0.27$ & $0.19$ & $0.24$ & $0.20$ & $0.28$ \\
        edge count & $0.07$ & $0.06$ & $0.15$ & $0.11$ & $0.26$ & $0.11$ & $0.14$ & $0.04$ & $0.11$ & $0.04$ \\
        has cycles & $0.44$ & $0.48$ & $0.58$ & $0.53$ & $0.49$ & $0.49$ & $0.61$ & $0.36$ & $0.37$ & $0.44$ \\
        is connected & $0.55$ & $0.54$ & $0.67$ & $0.64$ & $0.76$ & $0.59$ & $0.59$ & $0.22$ & $0.57$ & $0.56$ \\
        is tree & $0.49$ & $0.55$ & $0.57$ & $0.53$ & $0.50$ & $0.32$ & $0.48$ & $0.44$ & $0.49$ & $0.58$ \\
        max degree & $0.19$ & $0.14$ & $0.46$ & $0.33$ & $0.43$ & $0.28$ & $0.35$ & $0.17$ & $0.37$ & $0.20$ \\
        min degree & $0.24$ & $0.17$ & $0.58$ & $0.37$ & $0.49$ & $0.25$ & $0.54$ & $0.19$ & $0.65$ & $0.34$ \\
        node count & $0.35$ & $0.12$ & $0.63$ & $0.32$ & $0.34$ & $0.15$ & $0.55$ & $0.19$ & $0.38$ & $0.19$ \\
        \midrule
        full output & --- & $0.05$ & --- & $0.11$ & --- & $0.19$ & --- & $0.02$ & --- & $0.01$ \\
        \bottomrule
        \end{tabular}
    \end{subtable}
\end{table*}

\end{document}